\pdfoutput=1
\documentclass{article}

\PassOptionsToPackage{table}{xcolor}

\let\paperaddcontentsline\addcontentsline
\usepackage{iclr2027_conference,times}
\let\addcontentsline\paperaddcontentsline

\usepackage[utf8]{inputenc} %
\usepackage[T1]{fontenc}    %
\usepackage{xcolor}         %
\usepackage{graphicx}
\usepackage{amsmath}
\usepackage{amsfonts}       %
\usepackage{bm}
\usepackage{nicefrac}       %
\usepackage{booktabs}       %
\usepackage{xstring}        %

\newcommand{\cdbar}[1]{\textcolor[HTML]{#1}{\rule[-0.75ex]{1.25pt}{2.75ex}}}
\DeclareRobustCommand{\cdgroups}[1]{%
    \IfSubStr{#1}{a}{\cdbar{FFC857}}{\cdbar{FFFFFF}}\hspace{1.5pt}%
    \IfSubStr{#1}{b}{\cdbar{E9724C}}{\cdbar{FFFFFF}}\hspace{1.5pt}%
    \IfSubStr{#1}{c}{\cdbar{C5283D}}{\cdbar{FFFFFF}}\hspace{1.5pt}%
    \IfSubStr{#1}{d}{\cdbar{481D24}}{\cdbar{FFFFFF}}\hspace{1.5pt}%
    \IfSubStr{#1}{e}{\cdbar{255F85}}{\cdbar{FFFFFF}}%
}

\usepackage{algorithm}      %
\usepackage{algpseudocode}  %
\usepackage{microtype}      %
\usepackage{etoc}           %
\usepackage{url}            %
\usepackage{hyperref}       %

\title{Adapting Linear-Time Architectures\\ for Tabular In-Context Learning}

\definecolor{cornflowerblue}{rgb}{0.34, 0.51, 0.82}
\hypersetup{
    colorlinks=true, %
    linkcolor=magenta,          %
    citecolor=cornflowerblue,   %
    filecolor=magenta,          %
    urlcolor=magenta,           %
    pdftitle={Adapting Linear-Time Architectures for Tabular In-Context Learning},
    pdfauthor={David Schnurr, Felix Sarnthein, Thomas Hofmann, Imanol Schlag},
    }

\author{%
David Schnurr$^{1}$\quad
Felix Sarnthein$^{1,2,3}$\quad
Thomas Hofmann$^{1,4}$\quad
Imanol Schlag$^{1,4}$\\
\small $^{1}$ETH Z\"urich\quad
$^{2}$ELLIS Institute T\"ubingen\quad
$^{3}$MPI-IS\quad
$^{4}$ETH AI Center\\
\small Correspondence to \texttt{davidlschnurr@gmail.com}
}

\iclrfinalcopy
\begin{document}

\maketitle
\lhead{Preprint}

\begin{abstract}
Tabular foundation models achieve strong performance by conditioning on labelled examples in context, but softmax attention limits their use on large datasets. Existing linear-time alternatives, however, are mostly causal, and their potential for tabular in-context learning (ICL) remains underexplored. To address this, we (1) revisit causal training setups, (2) compare linear sequence mixers, and (3) investigate their ICL generalisation beyond the pretraining context length. First, we show that the best training setup for causal models resembles next-token prediction. Then, perhaps surprisingly, the most promising linear sequence mixer is causal: DeltaNet outperforms even non-causal linear attention. However, it degrades beyond $2$-$4\times$ the pretraining context length, and existing mitigation strategies such as bidirectionality defer the problem at best. A hidden-state oracle shows that this is not a capacity problem. Instead, our analysis points to an instability in the recurrent state, which drifts in deeper layers of causal models. Since DeltaNet's learned write rates overfit to the pretraining regime, we modulate them with a time-dependent decay schedule intervention to stabilise length generalisation. Finally, re-introducing non-causality by reading out from the final state allows us to closely match a controlled softmax attention baseline on OpenML-CC18 and TabArena.
\end{abstract}

\section{Introduction}

Despite the success of deep learning methods on text and images, gradient-boosted decision trees have long remained state-of-the-art for tabular data \citep{grinsztajn2022tree}. Prior-data fitted networks (PFNs) \citep{muller2022pfn} challenge this by recasting supervised learning as in-context prediction from a labelled context set. Models in this family, including TabPFN \citep{hollmann2023tabpfn,hollmann2025tabpfn}, TabICL \citep{jingangtabicl}, and TabDPT \citep{ma2025tabdptscalingtabularfoundation}, now match or exceed boosted-tree baselines without task-specific training \citep{erickson2026tabarena}.

Most of these so-called tabular foundation models rely on softmax self-attention \citep{vaswani2017attention}. 
Internally, predictions are non-parametric and require pairwise token interactions with all in-context training samples. 
Particularly for large datasets, linear-time architectures could present an attractive alternative as they compress the context into a fixed-size parametric state.
This potential has been confirmed by pioneering works which replaced softmax attention by linear-time sequence mixers \citep{zeng2025tabflex, kochstate, song2026feat, baur2024exploration}, but the design of the core memory mechanism has not been disentangled from other training and architectural decisions.
Modern linear recurrences such as DeltaNet update a fast-weight memory from past key-value pairs \citep{schlag2021lineartransformerssecretlyfast}, which can be interpreted as online optimisation to compress the observed context \citep{wang2025testtimeregression}. For tabular ICL, however, the fundamental question is if and how such potentially ordered updates can aggregate information across exchangeable rows in tables.

To that end, we adapt recurrent sequence mixers to tabular ICL rather than treating them as drop-in replacements for softmax attention. We perform a controlled study to disentangle training design, information flow, and update rules using matched parameter counts, a shared synthetic prior, and evaluation across the benchmarks OpenML-CC18 \citep{bischl2017openml} and TabArena \citep{erickson2026tabarena}. To quantify the generality of the learned ICL algorithm, we analyse models beyond the pretraining length and characterise novel failure mechanisms. %
Finally, we show how to adapt the ordered delta-rule to improve long-context stability and outperform order-invariant linear-time baselines, even on order-invariant tabular data.
Our contributions are:

\begin{itemize}
    \item \textbf{Training design space.} We identify label embedding and target formulation as two training axes used across prior work, and evaluate the resulting four strategies in Figure~\ref{fig:training_strategies} under a unified protocol. Interleaved Multi-Target (Int-MT) emerges as the strongest training strategy for causal and Combined Single-Target (Comb-ST) for non-causal models.
    \item \textbf{Controlled recurrence comparison.} We compare linear attention, Gated Linear Attention, DeltaNet, Gated DeltaNet, and Mamba-2 at matched scale. Delta-rule updates outperform other causal recurrences as well as non-causal linear attention at moderate sequence lengths.
    \item \textbf{Length-generalisation failure.} All recurrent variants degrade beyond roughly $2$-$4\times$ the pretraining context length, while order-invariant models remain stable. Common mitigations, including bidirectionality, state weaving \citep{moroshan2025tempopfn}, mimetic initialisation \citep{trockman2025mimetic}, and state passing \citep{buitrago2025understanding}, do not solve the issue.
    \item \textbf{Failure mechanisms and stabilisation.} A hidden-state oracle shows that fixed-size state capacity is not the bottleneck. Instead, causal degradation is accompanied by directional state drift in deeper layers. In DeltaNet we additionally identify overfitting of the write-rate $\beta_t$ to the sequence length and mitigate it by introducing an explicit decay term to $\beta_t$.
    \item \textbf{Final-State DeltaNet.} To remove the readout-level causal bottleneck, we propose a cheap but effective extension where every query reads directly from the stabilised final state.
\end{itemize}
Code is available at \url{https://github.com/schnurrd/ICL-Architectures}.

\section{Background}

Prior-data fitted networks (PFNs) \citep{muller2022pfn} cast supervised learning as conditional prediction from a context set. Given labelled context examples and unlabelled query points, a PFN predicts query labels in a single forward pass, without fitting task-specific parameters at test time. TabPFN \citep{hollmann2023tabpfn} applies this idea to tabular classification.

We consider multi-class classification datasets mapping $n$ inputs of $m$ features to $K$ classes, $\mathcal{D} = \{(x_i, y_i)\}_{i=1}^{n}$ with $x_i \in \mathbb{R}^m$ and $y_i \in \mathcal{Y} = \{1, \dots, K\}$, split into a labelled context set $\mathcal{C}$ of $n_c$ rows and a query set $\mathcal{Q}$ of $n_q$ rows. Then, a tabular foundation model $f_\theta$ maps the labelled context set $\mathcal{C}$ and an unlabelled query sample $x_i$ to a predictive class distribution $\hat p_{\theta,i} = f_\theta(\mathcal{C},x_i) \in \Delta^{K-1}$. 

During pretraining, datasets are sampled from a meta-distribution $p(\mathcal{D})$ over synthetic or real datasets, and the model is trained to predict held-out targets using the cross-entropy (CE) loss, i.e.
\begin{equation}
    \mathcal{L}(\theta) = \mathbb{E}_{\mathcal{D}\sim p(\mathcal{D})}\mathbb{E}_{(\mathcal{C}, \mathcal{Q}) \sim \operatorname{Split}(\mathcal{D})} 
    \left[
        \frac{1}{n_q} \sum_{(x_i, y_i) \in \mathcal{Q}} \mathrm{CE}(f_{\theta}(\mathcal{C}, x_i), y_i)
    \right] .
\end{equation}
As illustrated in Figure~\ref{fig:training_strategies}, query labels are typically hidden from the input context using a learned placeholder embedding $y_{\mathrm{pad}}$, but causal visibility could also be used as we explore in this paper. 

\subsection{Scaling Bottleneck of Transformers}

Most tabular foundation models use Transformer backbones \citep{vaswani2017attention}. This means that the model internally relies on non-parametric estimation using softmax attention to compute pairwise interactions between samples, similar to nearest-neighbour methods. Processing a context dataset $\mathcal{C}$ thus requires quadratic computation in the number of samples $n_c$. %
Predicting $\hat{p}_{\theta,i}$ from a single query sample $x_i$ still requires loading and iterating over the KV-cache of all $n_c$ samples. These scaling properties limit the applicability of attention-based models to large datasets. Linear-time architectures, in contrast, promise to compress the context $\mathcal{C}$ into a parametric state $\bm{S}_{n_c}$ and predict in constant time and memory (see Appendix~\ref{sec:scaling_analysis}), but it remains unclear if a generalising state exists and how to construct it. \looseness=-1 %

\subsection{Linear Attention and Recurrent Architectures}

Linear attention \citep{katharopoulos2020transformers} replaces the softmax kernel with a feature map $\phi(\cdot)$, s.t.
\begin{align}\label{eq:linear-attention}
    \bm{o}_i =
    \frac{1}{Z_i} \sum_{j \in \mathcal{V}(i)}\phi(\bm{q}_i)^\top \phi(\bm{k}_j) \bm{v}_j = \frac{1}{Z_i}\bm{S}_{\mathcal{V}(i)} \phi(\bm{q}_i), && \bm{S}_{\mathcal{V}(i)} = \sum_{j \in \mathcal{V}(i)}\bm{v}_j \phi(\bm{k}_j)^\top,
\end{align}
where $\bm{q}_i, \bm{k}_j \in \mathbb{R}^{d_{qk}}$, $\bm{v}_j\in\mathbb{R}^{d_v}$ are per-head queries, keys, and values obtained by linear projection of the token embeddings, $\mathcal{V}(i)$ denotes the set of visible tokens, and $Z_i$ is a normalisation term.
For non-causal attention, all queries process the same fixed-size state $\bm{S}_{n_c}=\bm{V}_{\mathcal{C}}^\top \phi(\bm{K}_{\mathcal{C}}) \in \mathbb{R}^{d_v \times d_{qk}}$ as a compressed key-value cache with $\bm{V}_{\mathcal{C}} \in \mathbb{R}^{n_c \times d_v}$ and $\bm{K}_{\mathcal{C}} \in \mathbb{R}^{n_c \times d_{qk}}$ stacking the context values and keys and $\phi$ applied row-wise. For causal linear attention the state evolves as the linear recurrence \[\bm{S}_i = \bm{S}_{i-1} + \bm{v}_i\phi(\bm{k}_i)^\top.\]

Modern linear-time recurrences mostly differ in how the memory state is updated. We focus on the most common update rules: Gated Linear Attention (GLA) \citep{yang2023gated} adds an input-dependent forget gate, DeltaNet \citep{schlag2021lineartransformerssecretlyfast} uses an overwriting mechanism based on the delta-rule, Gated DeltaNet \citep{yang2025gateddeltanetworksimproving} combines delta-rule updates with gating, and Mamba-2 \citep{dao2024transformers} admits a recurrent update through its state-space dual form. \looseness=-1 %

\begin{figure}[t]
  \centering
  \includegraphics[width=\textwidth,trim={18pt 12pt 0pt 0pt},clip]{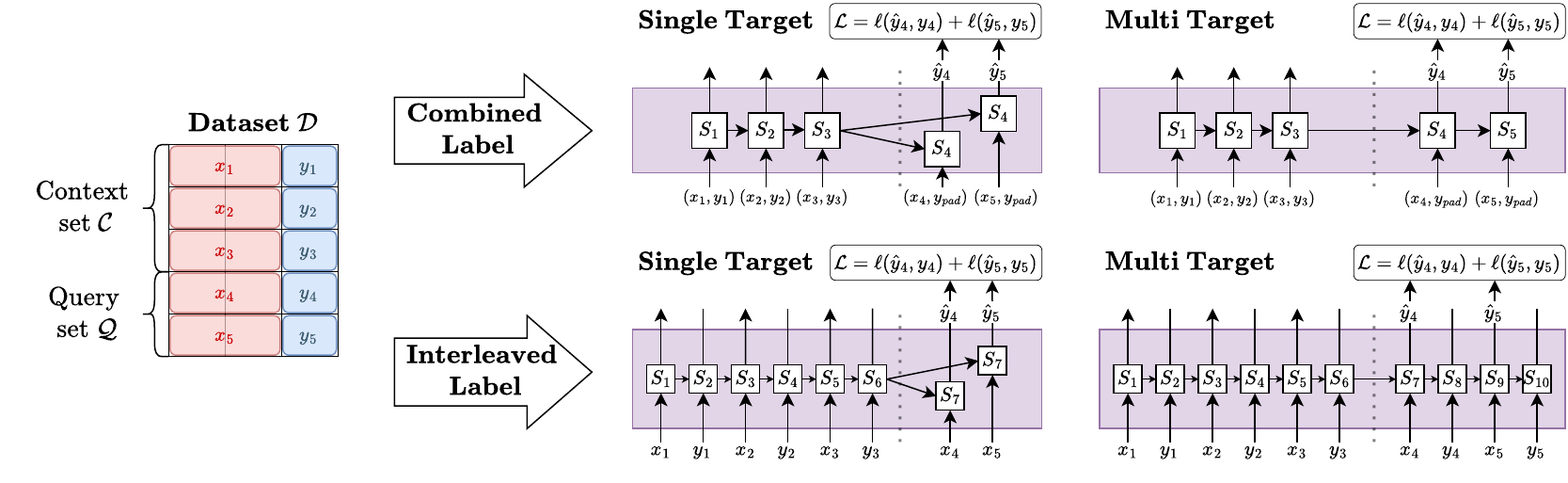}
  \caption{Training strategies given by label embedding and target formulation. Combined-label variants represent features and labels in one token; interleaved variants represent them separately. Single-target variants predict each query independently from the context; multi-target variants process target positions sequentially. Combined-label single-target is the standard (Tab)PFN setup.}
  \label{fig:training_strategies}
\end{figure}
\subsection{Causal Linear Recurrences as Layer-Local Test-Time Regressors}
\label{subsec:test_time_regressors}

The update rules above admit a common interpretation: the state is updated online from past key-value pairs. More broadly, related test-time regression and test-time training frameworks interpret sequence-model states as associative memories or weights adapted at inference time \citep{wang2025testtimeregression,sun2025TTT,behrouz2024titans,vonoswald2025mesanetsequencemodelinglocally}. The test-time regression (TTR) view \citep{wang2025testtimeregression} makes this connection explicit by interpreting the recurrent state as an online approximation to a layer-local associative-memory problem: %
\begin{equation}
\label{eq:ttr_objective}
    \bm{S}_t^\star = \operatorname*{argmin}_{\bm{S} \in \mathbb{R}^{d_v \times d_{qk}}} \frac{1}{2} \sum_{i=1}^t \lVert \bm{v}_i - \bm{S}\bm{k}_i \rVert_2^2 .
\end{equation}
In a related linear-regression analysis, causal in-context learners converge with increasing depth to potentially suboptimal stationary estimators corresponding to online gradient descent \citep{ding2023causallm}. For order-invariant tabular data, this motivates investigating whether the recurrent state converges to a stationary solution of (\ref{eq:ttr_objective}) as more samples are added to the context. We investigate this in Section~\ref{sec:understanding_generalization_failure}. \looseness=-1

\subsection{Tabular Foundation Models with Linear Backbones}

Recent work replaces the softmax backbone of tabular foundation models with linear-time alternatives. TabFlex \citep{zeng2025tabflex} swaps non-causal linear attention into TabPFN~v1. \citet{kochstate} replace transformer layers with Mamba \citep{gu2024mambalineartimesequencemodeling} and Hydra \citep{hydra} in the same backbone. \citet{baur2024exploration} explore autoregressive Mamba and Jamba \citep{lenz2025jamba} with interleaved feature--label tokens. FEAT \citep{song2026feat} stacks bidirectional Mamba with interleaved convolutional gated linear-attention layers for large structured data. Each work adopts a different combination of sequence mixer, label embedding, and target formulation, making it difficult to attribute gains to any single choice. Our study disentangles these axes and investigates length generalisation in order to quantify the generality of the learned ICL solution. Finally, we propose a principled solution to adapt the best-performing state construction mechanism to the tabular data setting. \looseness=-1 %

\section{Disentangling Training Strategies and Sequence Mixers}
\label{sec:adapting_causal_models_to_tabular_icl}

To capture the order-invariant nature of tabular data better, tabular models are usually non-causal with visibility $\mathcal{V}_{\mathrm{Non\text{-}Causal}}(i) = \{1, \dots, n_c\} \cup \{i\}$ and no positional encodings. In contrast, modern linear-time architectures are designed to be causal with visibility $\mathcal{V}_{\mathrm{Causal}}(i) = \{1, \dots, i\}$ and a strong positional bias. Reconciling the two seems almost impossible since even non-causal extensions are not order-invariant. While TabFlex~\citep{zeng2025tabflex} circumvents this by falling back to non-causal vanilla linear attention, we will now question the order-invariance assumption. Specifically, we first ask which training strategy is most effective for causal models (Section~\ref{subsec:Training_strategies}) and then which sequence mixers, including causal ones, perform best under matched parameter counts (Section~\ref{subsec:Architectural_Strategies}). \looseness=-1 %

\subsection{Pretraining and Evaluation Protocol}
\label{subsec:Evaluation_protocol}

To isolate how linear and recurrent models perform in tabular ICL, we keep the data-generating prior, training budget, evaluation protocol, and model sizes consistent across architectures. All models are pretrained exclusively on synthetic datasets sampled from the TabPFN v1 prior \citep{hollmann2023tabpfn}, containing at most $n=1000$ rows and $m=20$ input features unless stated otherwise. All architectures are matched to 12.5M parameters to ensure a fair comparison (deviations below $1\%$; details in Appendix~\ref{app:model_details}).
Most hyperparameters are reused from TabPFN v1, while backbone-specific hyperparameters are selected on validation performance on newly sampled synthetic datasets. 

We evaluate on the 30-dataset OpenML-CC18 \citep{bischl2017openml} subset of TabPFN v1, as well as the more modern TabArena classification datasets \citep{erickson2026tabarena}. All datasets are subsampled to at most $m=20$ features. OpenML-CC18 is further subsampled to at most $n=1000$ samples to match the pretraining size, while TabArena measures real-world length generalisation on larger datasets. We denote these preprocessed variants as OpenML-CC18$^*$ and TabArena$^*$. Following prior work, we report ROC-AUC and accuracy aggregated over 5-fold stratified CV splits and datasets.
For pairwise significance, we use paired Wilcoxon signed-rank tests \citep{wilcoxon1945individual} with Holm correction \citep{holm1979simple}. Further evaluation details are provided in Appendix~\ref{app:evaluation_details}.

\subsection{Training Strategies}
\label{subsec:Training_strategies}

It is not a priori clear whether design decisions made for tabular Transformers also transfer to causal variants. To that end, \citet{baur2024exploration} and \citet{kochstate} modify the training strategy but without a controlled comparison. %
We identify two axes, the label embedding and the target formulation. \looseness=-1

The {label embedding} determines row serialisation. The combined-label representation embeds features and labels into one token, while the interleaved representation embeds them in separate tokens: \looseness=-1
\begin{align}
    \text{combined:} \quad & \operatorname{emb}_i = \psi_x(x_i) + \psi_y(\tilde y_i), && L = n_c + n_q, \\[-1pt]
    \text{interleaved:} \quad & \operatorname{emb}_{2i-1} = \psi_x(x_i), \quad \operatorname{emb}_{2i} = \psi_y(y_i), && L=2(n_c + n_q),
\end{align}
where $\psi_x$ and $\psi_y$ are embedding functions, $\tilde y_i = y_i$ for context rows, and $\tilde y_i = y_{\operatorname{pad}}$ for query rows.
In the interleaved representation, labels are processed independently from feature tokens and therefore query labels can be held out with appropriate attention masking instead of input padding $y_{\mathrm{pad}}$.

The {target formulation} determines whether the query positions are treated independently or sequentially during pretraining. At the row level, single-target training isolates queries from each other, whereas multi-target training assumes ordering through causal visibility:
\begin{equation}
    \mathcal{V}_{\mathrm{ST}}(i) = \{1, \dots, n_c\} \cup \{i\}, \qquad 
    \mathcal{V}_{\mathrm{MT}}(i) = \{1,\dots,n_c\} \cup \{n_c+1,\dots,i\}.
\end{equation}
Recurrent architectures require careful adjustments to support single-target since all queries must condition on the same context state $\bm{S}_{n_c}$, while their predictions remain independent. %
We describe this ---to the best of our knowledge--- novel efficient implementation in Appendix~\ref{app:adding_stateless_prediction_to_fla}. 

Figure~\ref{fig:training_strategies} illustrates the resulting four strategies: $\textsc{Comb-ST}$ is typical for PFN-style softmax-attention models and TabFlex \citep{zeng2025tabflex}, \textsc{Comb-MT} is used by \citet{kochstate}, and \textsc{Int-MT}/\textsc{Int-ST} are used by \citet{baur2024exploration} for pretraining and prediction, respectively. To investigate the compatibility of causal linear sequence mixers with these training strategies, we aggregate the performance of four common models for each setup. We exclude standard linear attention because it cannot handle interleaved tokens without positional embeddings. To guarantee a fair comparison, we evaluate all models non-transductively, i.e. every queried test sample is predicted independently of the others as in single-target training.

The results in Table~\ref{tab:training_setup_performance} highlight that the training setup indeed needs to be considered jointly with the sequence mixer. Overall, Interleaved Multi-Target is most compatible with causal linear-time models.
We explain the strength of interleaving and multiple targets by a teacher-forcing effect similar to next-token prediction: feature-target pairs from the query set $\mathcal{Q}$ enter the context of later queries and effectively extend their context set $\mathcal{C}$ during pretraining. In comparison, interleaving with only single targets clearly deteriorates. 
For combined embeddings, however, multiple targets likely underperform due to a distribution shift because the placeholder labels $y_{\mathrm{pad}}$ can appear in the context set $\mathcal{C}$ during pretraining but not at inference. %
Finally, the Combined Single-Target setting remains very competitive considering that interleaving requires doubling the sequence length and thus the compute cost. It is also the only setting where causal models can be fairly compared to non-causal ones, and we therefore adopt the Comb-ST setting in the rest of this paper. We will, however, compare a causal Int-MT to a non-causal Comb-ST model in the final Table~\ref{tab:model_comparison_real_world}.

\begin{table}[t]
\vspace{-1em} %
\centering
\setlength{\tabcolsep}{2.5pt}
\renewcommand{\arraystretch}{1.0}
\caption{Aggregated training setup performance on OpenML-CC18$^*$ and TabArena$^*$, the preprocessed benchmark variants from Section~\ref{subsec:Evaluation_protocol}. Values are averaged across splits, datasets, and the four model families supporting all four setups: Gated Linear Attention, DeltaNet, Gated DeltaNet, and Mamba-2. Len. is the sequence length relative to combined-label models. Bold marks the best setup per metric. The vertical bars \cdgroups{abcde} denote insignificance classes per benchmark metric; setups sharing a bar are not significantly different under the paired Wilcoxon signed-rank test with Holm correction at $\alpha=5\%$.}
\label{tab:training_setup_performance}
\vspace{.8em}
\begin{tabular}{lccccc}
\toprule
&  & \multicolumn{2}{c}{\shortstack{OpenML-CC18$^*$}} & \multicolumn{2}{c}{\shortstack{TabArena$^*$}} \\
Training setup & Len.           & ROC-AUC $\uparrow$ & ACC $\uparrow$ %
                                & ROC-AUC $\uparrow$ & ACC $\uparrow$ %
                                \\
\midrule
Interleaved Multi-Target & $2\times$   & \textbf{0.8837} \cdgroups{a} & \textbf{0.8039} \cdgroups{a} %
                                & \textbf{0.8038} \cdgroups{a} & \textbf{0.8398} \cdgroups{a} %
                                \\
Combined Single-Target & $1\times$     & 0.8824 \cdgroups{b} & 0.8020 \cdgroups{a} %
                                & 0.8004 \cdgroups{b} & 0.8370 \cdgroups{b} %
                                \\
Interleaved Single-Target & $2\times$  & 0.8800 \cdgroups{c} & 0.7980 \cdgroups{b} %
                                & 0.8005 \cdgroups{b} & 0.8350 \cdgroups{c} %
                                \\
Combined Multi-Target & $1\times$      & 0.8793 \cdgroups{c} & 0.7952 \cdgroups{b} %
                                & 0.7978 \cdgroups{c} & 0.8338 \cdgroups{d} %
                                \\
\bottomrule
\end{tabular}

\end{table}

\subsection{Sequence Mixers}
\label{subsec:Architectural_Strategies}

Linear attention based architectures usually share the readout mechanism $\bm{o}_i = \bm{S}_i \bm{q}_i$ and differ in the state construction. The most common ingredients for recurrent update rules on top of vanilla linear attention $\bm{S}_t = \bm{S}_{t-1} + \bm{v}_t \bm{k}_t^{\top}$ \citep{katharopoulos2020transformers} are gating $\bm{S}_t = \bm{S}_{t-1}\mathrm{Diag}(\alpha_t) + \bm{v}_t \bm{k}_t^{\top}$ with a forgetting strength $\alpha_t$ \citep{yang2023gated}, the delta rule $\bm{S}_t = \bm{S}_{t-1}\big(\mathbf{I} - \beta_t \bm{k}_t \bm{k}_t^{\top}\big) + \beta_t \bm{v}_t \bm{k}_t^{\top}$ which reduces memory collisions based on a write strength $\beta_t$ \citep{schlag2021lineartransformerssecretlyfast}, or combinations thereof \citep{yang2025gateddeltanetworksimproving}.
Note that these updates are not order-invariant. To isolate the effect of this core memory mechanism, we compare these update rules in the Comb-ST setting. We further train a parameter-matched softmax attention baseline and report full-scale tabular baselines for context.

Table~\ref{tab:openml_tabarena_summary} shows that non-causal softmax attention is the strongest controlled architecture, confirming that simply replacing it with linear or recurrent updates reduces performance on both benchmarks. Among linear sequence mixers, the causal DeltaNet and its gated variant perform best overall. This indicates that delta-rule updates transfer better to tabular ICL than vanilla linear attention. Gating alone does not meaningfully change performance, and Mamba-2 trails the rest across all metrics. Within linear attention, the non-causal version only outperforms its causal counterpart significantly on OpenML-CC18 ROC-AUC. Together, these results suggest that the update rule of recurrent sequence mixers is more important than causality or order-invariance.

We want to highlight that causal DeltaNet and non-causal softmax attention remain competitive with non-causal TabFlex \citep{zeng2025tabflex} on OpenML-CC18 despite our far smaller pretraining budgets. On TabArena, however, DeltaNet deteriorates compared to the baselines. Recall that this benchmark includes datasets larger than our 1000-row pretraining cap to measure real-world length generalisation. Figure~\ref{fig:real_world_seqlen} in the appendix confirms that the gap to softmax attention generally grows with increasing dataset size. In Section~\ref{sec:understanding_generalization_failure} we will investigate this length-generalisation failure in more detail, relying on the synthetic prior to allow for finer context-length coverage than TabArena.

\begin{table}[t]
\vspace{-1em} %
\centering
\setlength{\tabcolsep}{4pt}
\renewcommand{\arraystretch}{1.0}
\caption{Model comparison on the preprocessed OpenML-CC18$^*$ and TabArena$^*$ benchmark variants (Section~\ref{subsec:Evaluation_protocol}). The upper block reports full-scale released baselines, not matched in parameter count or pretraining budget, while the lower block reports our controlled backbones, all using Combined Single-Target. Values are averaged over 5-fold stratified CV splits and datasets. Bold and underline mark the best and second-best controlled models per metric. The vertical bars \cdgroups{abcde} denote pairwise insignificance classes per metric; models sharing a bar are not significantly different.}
\label{tab:openml_tabarena_summary}
\vspace{0.8em}
\begin{tabular}{l c c c c }
\toprule
& \multicolumn{2}{c}{\shortstack{OpenML-CC18$^*$}} & \multicolumn{2}{c}{\shortstack{TabArena$^*$}} \\
                        & ROC-AUC $\uparrow$ & ACC $\uparrow$ %
                        & ROC-AUC $\uparrow$ & ACC $\uparrow$ %
                        \\
\midrule
TabPFN v2.5             & 0.8993 \cdgroups{} & 0.8298 \cdgroups{} %
                        & 0.8383 \cdgroups{} & 0.8642 \cdgroups{} %
                        \\
TabICLv2                & 0.8971 \cdgroups{} & 0.8286 \cdgroups{} %
                        & 0.8375 \cdgroups{} & 0.8631 \cdgroups{} %
                        \\
TabFlex                 & 0.8859 \cdgroups{} & 0.8093 \cdgroups{} %
                        & 0.8089 \cdgroups{} & 0.8463 \cdgroups{} %
                        \\
CatBoost                & 0.8833 \cdgroups{} & 0.8062 \cdgroups{} %
                        & 0.8231 \cdgroups{} & 0.8562 \cdgroups{} %
                        \\
\midrule
Softmax Non-Causal  & \textbf{0.8900} \cdgroups{a} & \textbf{0.8166} \cdgroups{a} %
                        & \textbf{0.8172} \cdgroups{a} & \textbf{0.8506} \cdgroups{a} %
                        \\      
Delta              & \underline{0.8860} \cdgroups{abc} & \underline{0.8078} \cdgroups{ab} %
                        & 0.8074 \cdgroups{b} & 0.8388 \cdgroups{bc} %
                        \\    
Delta Gated       & 0.8857 \cdgroups{bdc}  & 0.8067 \cdgroups{bc} %
                        & \underline{0.8078} \cdgroups{b} & \underline{0.8414} \cdgroups{b} %
                        \\   
Linear Non-Causal   & 0.8833 \cdgroups{bd} & 0.8039 \cdgroups{b} %
                        & 0.8004 \cdgroups{c} & 0.8396 \cdgroups{bc} %
                        \\
Linear Gated   & 0.8827 \cdgroups{de} & 0.7985 \cdgroups{c} %
                        & 0.8038 \cdgroups{c} & 0.8388 \cdgroups{bc} %
                        \\
Linear Causal   & 0.8817 \cdgroups{ce}  & 0.7988 \cdgroups{bc}  %
                        & 0.8033 \cdgroups{c}  & 0.8390 \cdgroups{bc}  %
                        \\
Mamba-2                 & 0.8784 \cdgroups{e} & 0.7951 \cdgroups{c} %
                        & 0.7912 \cdgroups{d} & 0.8344 \cdgroups{c}  %
                        \\
\bottomrule
\end{tabular}

\vspace{-0.0em}
\end{table}

\section{Understanding Generalisation Failure}
\label{sec:understanding_generalization_failure}

We now analyse why the competitive short-context performance of causal models does not generalise to larger context sets. %
To that end, we sample 500 synthetic datasets of size 128k from which we construct query sets with fixed $n_q=100$ queries and controlled context sets of increasing size $n_c$.

\subsection{Length Generalisation Gap and Hidden-State Capacity} \label{subsec:length_generalisation_and_state_capacity}

\begin{figure}[t]
    \centering
    \includegraphics[width=\textwidth, trim={0pt, 32pt, 0pt 8pt}, clip]{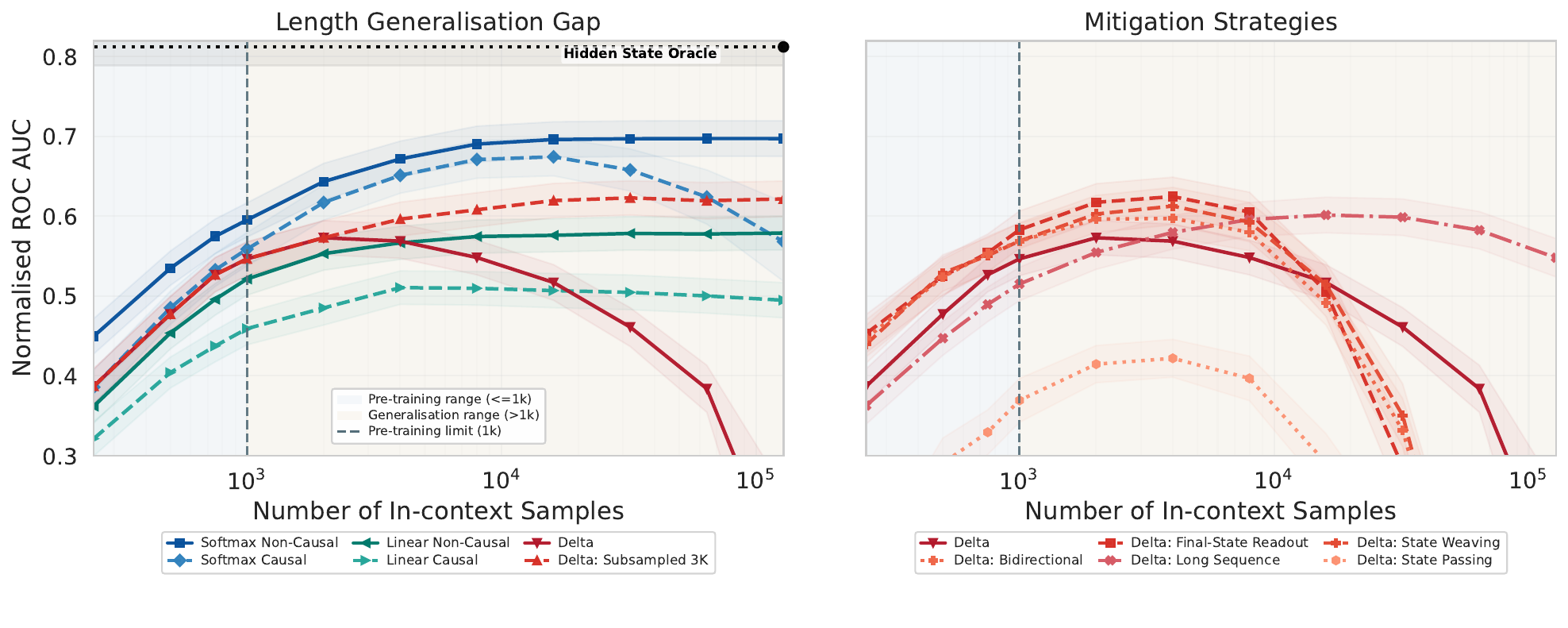}
    \caption{Length generalisation beyond the 1k pretraining length. We report normalised ROC-AUC averaged over 500 synthetic datasets with 95\% confidence intervals.
    \textbf{(a) Length Generalisation Gap:} Causal models underperform their non-causal counterparts. DeltaNet is initially competitive but degrades more strongly, while subsampled DeltaNet remains stable; the oracle is evaluated only at 128k and shown as a horizontal reference.
    \textbf{(b) Mitigation Strategies:} None of the applicable strategies eliminates DeltaNet's long-context degradation (Section~\ref{subsec:Mitigation_strategies}).}
    \label{fig:causal_and_oracle_seq_len_comparison}
    \vspace{-1em}
\end{figure}

Figure~\ref{fig:causal_and_oracle_seq_len_comparison}(a) compares the causal and non-causal variants of softmax and linear attention beyond the 1000-row pretraining regime. Consistent with \citet{zeng2025tabflex}, causal models degrade as the number of samples increases, from $2$-$4\times$ the pretraining range, with severity varying by update rule: causal additive linear attention degrades only moderately, while more expressive rules such as DeltaNet and GLA degrade more sharply. In contrast, non-causal attention continues to generalise, suggesting that causal prefix access is one important source of the failure as we will analyse further in Section~\ref{sec:hidden_state_drift_causal_linear_attention}. \looseness=-1

To test whether additional samples could improve prediction, we partition the context into blocks of 3K samples and ensemble predictions for each query across the blocks. This protocol already allows DeltaNet to consistently outperform non-causal linear attention. It shows that DeltaNet could benefit from additional samples, but fails to effectively aggregate them into states $\bm{S}_{n_c}$ over long sequences. \looseness=-1

Still, it remains open whether the fixed-size hidden state even provides the capacity to exploit additional samples compared to softmax attention. To test this hypothesis, we design a hidden-state oracle for the pretrained and frozen DeltaNet checkpoint. We then optimise only its hidden-states $\bm{S}$ with gradient descent such that the end-to-end model fits the 128k context samples %
(details in Appendix~\ref{app:Hidden_state_oracle_details}).
In Figure~\ref{fig:causal_and_oracle_seq_len_comparison}(a), the oracle achieves strong long-context performance, demonstrating that limited memory capacity is not the bottleneck: there exists a parametric state $\bm{S}$ which even outperforms our non-parametric softmax attention baseline, but existing linear models do not construct it.\looseness=-1

We use this end-to-end oracle as a diagnostic tool, but it would be valid under the non-transductive evaluation protocol. In Appendix~\ref{app:real_data_oracle} we show that performance transfers to the five largest TabArena datasets. Since hidden-state optimisation requires careful length-dependent tuning, we leave practical deployment to future work. Approximating the end-to-end test-time objective with the layer-local regression objective of Equation~\ref{eq:ttr_objective} further allows computing an optimal state $\bm{S}^{\mathrm{ridge}}$ from a permutation-invariant, closed-form ridge regression. In Appendix~\ref{app:layer_local_ridge_comparison} we show that such a sequence mixer can be trained and generalises, supporting the hypothesis that stable state construction is the key challenge.

\subsection{Mitigation Strategies}
\label{subsec:Mitigation_strategies}

We evaluate several mitigation strategies for causal recurrent models. \textit{Bidirectionality} \citep{afzal2026linear} processes the context in both directions and writes to the same hidden states $\bm{S}_i$. Our \textit{final-state readout} modification (Section~\ref{subsec:final_state_readout}) reads from the same final context state $\bm{S}_{n_c}$ at all positions $\bm{q}_i$, allowing for non-causal information flow through the aggregated final state. \textit{State weaving} \citep{moroshan2025tempopfn} initialises each layer with the final hidden state of the previous layer to introduce non-causality. \textit{State passing} \citep{buitrago2025understanding} reuses cached recurrent states from prior batches during pretraining. \textit{Long-sequence pretraining} biases the row sampler toward longer contexts while remaining compute-matched by sampling smaller datasets more frequently. %

Figure~\ref{fig:causal_and_oracle_seq_len_comparison}(b) shows all applicable mitigations for DeltaNet; results for GLA are in Appendix~\ref{app:Generalisation_and_mitigation_strategies_for_gla}. Overall, none of these strategies provides a general solution. Bidirectionality, final-state readout, and state weaving improve performance near the pretraining range, indicating that non-causal information flow is important. However, they do not solve length generalisation. While non-causality stabilises vanilla linear attention, it increases degradation in DeltaNet. Length degradation is therefore not fully explained by causal access alone: a hidden state may itself be an unstable estimator of the task, even when all positions write to or read from it. Long-sequence pretraining shifts degradation to longer contexts at the cost of short-context performance, and state passing is recurrence-dependent: it reduces degradation for GLA, but lowers overall performance for DeltaNet.

\subsection{Hidden-State Drift in Causal Linear Attention}
\label{sec:hidden_state_drift_causal_linear_attention}

We next analyse the recurrent hidden states as a function of sequence length, focusing on vanilla causal linear attention as the simplest example reproducing the failure and comparing it with non-causal linear attention, which generalises in our setting. Neither model uses positional encodings or the normalisation $\frac{1}{Z_i}$ from Equation~\ref{eq:linear-attention}. Visualisations are provided in Appendix~\ref{app:Generalisation_Failure_Analysis}. Comparing the Frobenius norm of the final state $\bm{S}_{n_c}$ for increasing context sizes $n_c$ (Figure~\ref{fig:hidden_state_debug_frobenius_unnormalised}, second panel), the causal model's hidden-state norms drift substantially relative to one another across layers as the context grows, whereas non-causal linear attention exhibits approximately parallel norm growth. If magnitude drift were the cause, controlling the state norm should restore performance. Training with Frobenius normalisation of each layer's state matrix does not remove the degradation (Figure~\ref{fig:hidden_state_debug_frobenius_unnormalised}, first panel), so magnitude drift is not sufficient to explain the failure.

We then measure directional drift by comparing each normalised hidden state $\bm{S}_{n_c}$ to the corresponding state $\bm{S}_{1000}$ on the same dataset at pretraining length $n_c=1000$, using cosine similarity after flattening the state matrix. Figure~\ref{fig:hidden_state_debug_frobenius_unnormalised} (bottom panel) shows that non-causal linear attention approaches a stable hidden state, whereas causal linear attention drifts increasingly far from $\bm{S}_{1000}$, with performance degradation occurring alongside the drift. The drift appears in earlier layers and becomes substantially stronger in deeper layers, consistent with accumulating deviations across successive causal layers. Similar drift appears across different data-generating priors whenever the pretraining distribution is sufficiently diverse, the target multi-class, and the model is sufficiently deep. This suggests a failure mode beyond the online-gradient-descent interpretation of causal linear self-attention \citep{ding2023causallm}, which does not account for the persistent directional drift we observe.\footnote{
We provide a minimal notebook for synthetic pretraining and sequence-length evaluation that reproduces this failure mode using a simple MLP prior, Gaussian features, and uniformly distributed class labels.}

\subsection{DeltaNet Length Overfitting and Write-Rate Decay}
\label{subsec:recurrent_state-estimator}

\begin{figure}[t]
    \centering
    \includegraphics[width=\textwidth, trim={3pt 0pt 3pt 0pt}, clip]{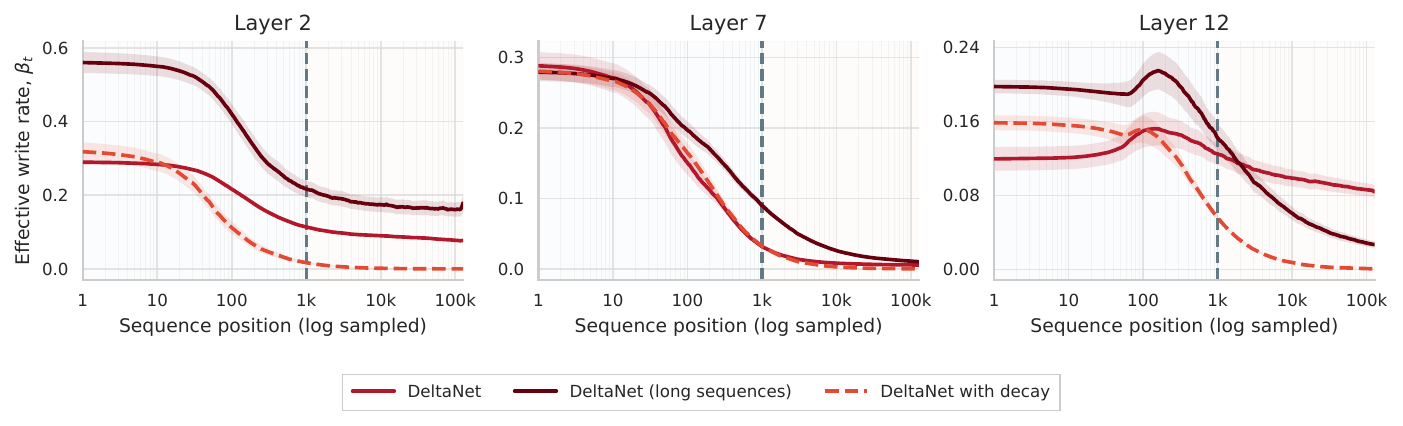}
    \caption{Effective write rates $\beta_t^\mathrm{eff}$ over sequence positions across network depth. We report means over 50 datasets, including the 95\% CIs. The dashed vertical line marks the 1000-row pretraining limit, with additional layer plots provided in Figure~\ref{fig:deltanet_beta_over_seq_len_full} of the appendix. Standard DeltaNet learns a decreasing write-rate profile, but compared to the (not compute-matched) long-sequence pretrained variant, it lowers the write rates less aggressively at very deep layers. The decay variant is our proposed solution, which imposes the decay explicitly within the recurrence relation.}
\label{fig:deltanet_beta_over_seq_len}
\vspace{-0.8em}
\end{figure}

The hidden-state drift alone does not explain why causal additive linear attention degrades only moderately in Figure~\ref{fig:causal_and_oracle_seq_len_comparison}(a) while DeltaNet degrades much more sharply. Note that the recurrence of additive linear attention forms an order-invariant set aggregate for fixed layer-local keys and values and corresponds to a single gradient step on the objective of Equation~\ref{eq:ttr_objective}. The delta-rule, on the other hand, forms an ordered aggregate of first-order updates on per-sample losses \citep{wang2025testtimeregression}. Its endpoint $\bm{S}_{n_c}$ is therefore the result of an ordered online optimisation trajectory controlled by the update rule $\bm{S}_t = \bm{S}_{t-1}\big(\mathbf{I} - \beta_t \bm{k}_t \bm{k}_t^{\top}\big) + \beta_t \bm{v}_t \bm{k}_t^{\top}$ with its learned write rates $\beta_t$.

Classical stochastic approximation motivates decaying the effective step sizes to achieve stable long-horizon convergence \citep{robbins1951stochastic}. In contrast, DeltaNet learns its write-rate schedule from pretraining on sequences of at most 1000 rows. Figure~\ref{fig:deltanet_beta_over_seq_len} depicts the learned write rates: standard DeltaNet learns a decreasing schedule by inferring position from the evolving causal representations. In deeper layers, the long-context pretrained variant lowers its late-sequence write rates substantially, coinciding with better long-context performance. Overall, the figure indicates that DeltaNet overfits its update dynamics to the pretraining length, resulting in excessively large updates at long sequence lengths, which motivates a concrete intervention. We introduce a write-rate decay to explicitly decay the effective write rate over time by setting $\beta_t^\mathrm{eff} = \beta_t\eta_t$ with $\eta_t = c/(t+c)$, where $c=256$ is selected on synthetic pretraining performance. With this decay, DeltaNet continues to improve beyond the pretraining length instead of collapsing and exhibits only minor degradation at long sequence lengths (full analysis in Appendix~\ref{app:deltanet_update_dynamics}).

\section{Final-State Readout DeltaNet with Write-Rate Decay}
\label{subsec:final_state_readout}

Having stabilised DeltaNet for long contexts, we now introduce our non-causal readout extension to the architecture. \textit{Final-state readout} enables non-causal information flow by reading from the final state $\bm{o}_i = \bm{S}_{n_c}\bm{q}_i$ instead of $\bm{o}_i = \bm{S}_i\bm{q}_i$. Unlike the bidirectional approach, this requires only a single forward pass, and it outperforms all other non-causal variants within the stable range in Figure~\ref{fig:causal_and_oracle_seq_len_comparison}(b). The $\beta_t$ decay again stabilises the final state $\bm{S}_{n_c}$, enables length generalisation, and improves performance compared to causal DeltaNet on the synthetic tasks as shown in Figure~\ref{fig:standard_and_final_state_deltanet_with_decay_intervention} in the appendix. \looseness=-1

\begin{table}[t]
\vspace{-1em} %
\centering
\setlength{\tabcolsep}{4pt}
\renewcommand{\arraystretch}{0.97}
\caption{Benchmark performance of the DeltaNet variants and the non-causal attention baselines (protocol of Section~\ref{subsec:Evaluation_protocol}). FS denotes final-state readout, Decay denotes the time-dependent decay intervention, and Int-MT denotes the Interleaved Multi-Target configuration (Appendix~\ref{app:Pre-training_and_Hyperparameter_details}). All other models use Combined Single-Target. Bold and underline mark the best and second-best models per metric. Significance bars \cdgroups{abcde} mark non-significant differences between models within one group.}
\label{tab:model_comparison_real_world}
\vspace{0.8em}
\begin{tabular}{lcccc}
\toprule
& \multicolumn{2}{c}{OpenML-CC18$^*$} & \multicolumn{2}{c}{TabArena$^*$} \\
& ROC-AUC $\uparrow$ & ACC $\uparrow$ & ROC-AUC $\uparrow$ & ACC $\uparrow$ \\
\midrule
Softmax Non-Causal & \textbf{0.8900} \cdgroups{a} & \textbf{0.8166} \cdgroups{ab} & \textbf{0.8172} \cdgroups{a} & \textbf{0.8506} \cdgroups{ac} \\
Delta-FS + Decay & \underline{0.8884} \cdgroups{ab} & \underline{0.8145} \cdgroups{ab} & 0.8139 \cdgroups{a} & \underline{0.8468} \cdgroups{a} \\
Delta + Decay (Int-MT) & 0.8873 \cdgroups{ab} & 0.8120 \cdgroups{ab} & \underline{0.8141} \cdgroups{a} & 0.8459 \cdgroups{ac} \\
Delta + Decay & 0.8864 \cdgroups{ab} & 0.8091 \cdgroups{ab} & 0.8100 \cdgroups{b} & 0.8438 \cdgroups{bc} \\
Delta-FS & 0.8874 \cdgroups{ab} & 0.8137 \cdgroups{a} & 0.8022 \cdgroups{bcd} & 0.8352 \cdgroups{ab} \\
Delta & 0.8860 \cdgroups{ab} & 0.8078 \cdgroups{b} & 0.8074 \cdgroups{c} & 0.8388 \cdgroups{b} \\
Linear Non-Causal & 0.8833 \cdgroups{b} & 0.8039 \cdgroups{b} & 0.8004 \cdgroups{d} & 0.8396 \cdgroups{b} \\
\bottomrule
\end{tabular}

\end{table}

Table~\ref{tab:model_comparison_real_world} evaluates the resulting models on real-world tasks following Section~\ref{subsec:Evaluation_protocol}.
DeltaNet with final-state readout and write-rate decay is the strongest evaluated linear-time variant: it improves over standard DeltaNet on both benchmarks. It slightly outperforms the causal Delta + Decay in the Interleaved Multi-Target setting on three of four metrics, while processing half as many tokens per dataset. Appendix~\ref{app:extended_model_comparison} further reports an extended comparison and shows that it aggregates information even better than the subsampled 3K and layer-local ridge regression baselines of Section~\ref{subsec:length_generalisation_and_state_capacity}. Finally, it substantially closes the significance gap to non-causal softmax attention on TabArena, which indicates that the improved length generalisation transfers to real-world datasets.

\section{Discussion}
\label{sec:discussion}
In this study, we first disentangled the interaction of training design, information flow, and update rule. We show that Interleaved Multi-Target is the most suitable setting for causal models, but requires $2\times$ sequence length and is not valid for non-causal models. When compared in the more standard Combined Single-Target setting, we observe that the nature of the update rule is more important than non-causal information flow: at moderate context lengths, causal DeltaNet variants outperform our non-causal linear attention baseline and remain competitive with the externally trained TabFlex checkpoint. We then identify length generalisation as the main challenge for such order-dependent update rules even when they are extended to support non-causal information flow.%

The hidden-state oracle rules out state capacity as the bottleneck for length generalisation. Instead, the recurrent update dynamics drift when the context is extended beyond the pretraining range. Under the test-time regression view, DeltaNet overfits its write rates $\beta_t$ to the pretraining length rather than converging to a stable estimator. To mitigate this, we propose a time-dependent $\beta_t$ decay. The resulting model is the best-performing linear-time model in our evaluation, closing most of the gap to softmax attention, especially in the Int-MT setting or when extended for non-causality via final-state readout. \looseness=-1

\textbf{Limitations and future work.} We prioritise a controlled comparison over absolute performance. Larger pretraining budgets and further tuning may disproportionately improve performance. We use row-wise tabular ICL, but expect the results to extend to modern two-axis attention of TabPFN v2 \citep{hollmann2025tabpfn} or TabICL's token preprocessing \citep{jingangtabicl}. Future work could explain the hidden-state drift, build on the write-rate decay, or develop update rules that are length-stable by design. Another interesting direction could aim to construct the (non-causal) oracle hidden state, for example via end-to-end test-time training \citep{tandon2025end}. %

\textbf{Conclusion.} Through the lens of test-time regression, linear architectures compress the rows of tabular datasets into a fixed-size parametric state, either layer-locally or end-to-end. This paper shows across various considered settings that stable state construction is the key challenge to learn a generalising ICL algorithm. To that end, we adapt ordered fast-weight updates in both causal and non-causal models to stably aggregate information across exchangeable rows in tabular data. We hope that these insights pave the way for scalable tabular foundation models based on parametric estimation. \looseness=-1

\subsection*{Acknowledgements}

This work was mainly conducted during David Schnurr's Master's thesis in the Data Analytics Lab at ETH Z\"urich. We thank the Data Analytics Lab for providing access to their computing resources. Felix Sarnthein acknowledges the financial support of the Hector Foundation and the Max Planck ETH Center for Learning Systems (CLS).

\subsection*{Reproducibility Statement}
\label{app:reproducibility}

\textbf{Code Availability.} The code for all experiments is available at
\url{https://github.com/schnurrd/ICL-Architectures}. Our implementation builds on the PFNs codebase of \citet{muller2022pfn}, available at \url{https://github.com/SamuelGabriel/PFNs}. We provide the full codebase, including the pretraining and evaluation pipelines, experiment configs, and notebooks. Most of the linear-time models in the codebase were imported from the Flash Linear Attention Library \citep{yang2024fla} and adapted to the tabular prediction setting with custom patches as described in Appendix~\ref{app:adding_stateless_prediction_to_fla}. We additionally provide a minimal notebook reproducing the length-generalisation failure mode of Section~\ref{sec:hidden_state_drift_causal_linear_attention} using a simple MLP prior, Gaussian features, and uniformly distributed class labels.

\textbf{Data Availability.} Synthetic pretraining and length-generalisation datasets are generated on the fly from the TabPFN v1 prior. All real-world datasets used in our experiments, specifically for the OpenML-CC18 \citep{bischl2017openml} and TabArena \citep{erickson2026tabarena} benchmark evaluation, are freely available at \url{https://www.openml.org} \citep{Vanschoren_2014} and are downloaded automatically when running the evaluation in our codebase.

\textbf{Pretraining and Hyperparameter Details.} An overview of the pretraining setup is described in Section~\ref{subsec:Evaluation_protocol} with additional information in Appendix~\ref{app:training_details}. 

\textbf{Evaluation Reproducibility.} The evaluation protocol is described in Section~\ref{subsec:Evaluation_protocol}, with additional details about metrics and the significance testing described in Appendix~\ref{app:evaluation_details}.

\subsection*{Ethics Statement}

This work is methodological and studies scalable architectures for tabular in-context learning. Potential positive impacts include more efficient use of tabular foundation models, reduced inference costs on large context sizes, and longer model lifetimes by enabling pretrained models to use more data at prediction time without task-specific retraining. Such improvements could make tabular foundation models accessible in low-compute or latency-constrained settings, including edge devices.

Potential negative impacts arise from the broader use of more capable tabular prediction systems. Efficient long-context architectures could increase the deployment of tabular prediction systems on large structured data collections, where they may be misused without appropriate validation and oversight. This might reinforce biases through automated decision-making. Additionally, while tabular foundation models can be deployed to new tasks efficiently without task-specific training, they require significant initial compute resources for pretraining. Our focus on linear architectures aims to reduce this cost for long-context pretraining and improve inference efficiency.

\subsection*{AI Use Statement}

We used generative artificial intelligence tools, including OpenAI's GPT, Anthropic's Claude, and Google's Gemini, in a supporting role during the writing and implementation phases of this project. 

During the writing phase, we used the tools for language editing, grammar checking, and sentence- and sometimes paragraph-level reformulation based on our initial drafts and then continued to refine them manually.
The literature search and review were carried out manually by us.
We used the tools only for additional checks to find further relevant work and checked every suggested paper before using it in any capacity.
During the implementation phase, we used the tools for coding assistance, such as implementing individual functions according to our specifications, wiring new parameters through the model code, helping write plotting and testing code, reviewing our hand-written code additions, and cleaning up implementations. Specifically, we designed and implemented the custom stateless prediction patch for single-target inference for linear attention first and then applied the patch to other models with tool assistance (Appendix~\ref{app:adding_stateless_prediction_to_fla}).

We want to be clear that we did not use AI to determine central parts such as the methodology, experimental design, interpretation of results, or conclusions. We also did not use the tools to define the research questions, make experimental decisions, interpret results, or draw scientific conclusions.
We reviewed all AI-assisted text edits and suggestions, reviewed and tested all AI-assisted code suggestions, and take full responsibility for the final content of this work.

\bibliographystyle{iclr2027_conference}

\bibliography{main}

@inproceedings{
    muller2022pfn,
    title={Transformers Can Do Bayesian Inference},
    author={Samuel M{\"u}ller and Noah Hollmann and Sebastian Pineda Arango and Josif Grabocka and Frank Hutter},
    booktitle={International Conference on Learning Representations},
    year={2022},
    url={https://openreview.net/forum?id=KSugKcbNf9}
}

@inproceedings{
  hollmann2023tabpfn,
  title={Tab{PFN}: A Transformer That Solves Small Tabular Classification Problems in a Second},
  author={Noah Hollmann and Samuel M{\"u}ller and Katharina Eggensperger and Frank Hutter},
  booktitle={The Eleventh International Conference on Learning Representations},
  year={2023},
  url={https://openreview.net/forum?id=cp5PvcI6w8_}
}

@article{hollmann2025tabpfn,
 title={Accurate predictions on small data with a tabular foundation model},
 author={Hollmann, Noah and M{\"u}ller, Samuel and Purucker, Lennart and
         Krishnakumar, Arjun and K{\"o}rfer, Max and Hoo, Shi Bin and
         Schirrmeister, Robin Tibor and Hutter, Frank},
 journal={Nature},
 year={2025},
 month={01},
 day={09},
 doi={10.1038/s41586-024-08328-6},
 publisher={Springer Nature},
 url={https://www.nature.com/articles/s41586-024-08328-6},
}

@inproceedings{jingangtabicl,
  title={Tab{ICL}: {A} Tabular Foundation Model for In-Context Learning on Large Data},
  author={Qu, Jingang and Holzm{\"u}ller, David and Varoquaux, Ga{\"e}l and Le Morvan, Marine},
  booktitle={International Conference on Machine Learning},
  year={2025}
}

@inproceedings{qu2026tabiclv2,
  title={{TabICLv2}: {A} better, faster, scalable, and open tabular foundation model},
  author={Qu, Jingang and Holzm{\"u}ller, David and Varoquaux, Ga{\"e}l and Le Morvan, Marine},
  booktitle={International Conference on Machine Learning},
  year={2026}
}

@inproceedings{
    zeng2025tabflex,
    title={TabFlex: Scaling Tabular Learning to Millions with Linear Attention},
    author={Yuchen Zeng and Tuan Dinh and Wonjun Kang and Andreas C Mueller},
    booktitle={Forty-second International Conference on Machine Learning},
    year={2025},
    url={https://openreview.net/forum?id=d60cmFf89H}
}

@inproceedings{
vonoswald2025mesanetsequencemodelinglocally,
title={MesaNet: Sequence Modeling by Locally Optimal Test-Time Training},
author={Johannes von Oswald and Nino Scherrer and Seijin Kobayashi and Luca Versari and Songlin Yang and Maximilian Schlegel and Kaitlin Maile and Yanick Schimpf and Oliver Sieberling and Alexander Meulemans and Guillaume Lajoie and Rif A. Saurous and Charlotte Frenkel and Razvan Pascanu and Blaise Aguera y Arcas and Joao Sacramento},
booktitle={The Fourteenth International Conference on Learning Representations},
year={2026},
url={https://openreview.net/forum?id=xa3OnTb6c3}
}

@article{tandon2025end,
  title={End-to-end test-time training for long context},
  author={Tandon, Arnuv and Dalal, Karan and Li, Xinhao and Koceja, Daniel and R{\o}d, Marcel and Buchanan, Sam and Wang, Xiaolong and Leskovec, Jure and Koyejo, Sanmi and Hashimoto, Tatsunori and others},
  journal={arXiv preprint arXiv:2512.23675},
  year={2025}
}

@inproceedings{
sun2025TTT,
title={Learning to (Learn at Test Time): {RNN}s with Expressive Hidden States},
author={Yu Sun and Xinhao Li and Karan Dalal and Jiarui Xu and Arjun Vikram and Genghan Zhang and Yann Dubois and Xinlei Chen and Xiaolong Wang and Sanmi Koyejo and Tatsunori Hashimoto and Carlos Guestrin},
booktitle={Forty-second International Conference on Machine Learning},
year={2025},
url={https://openreview.net/forum?id=wXfuOj9C7L}
}

@inproceedings{
behrouz2024titans,
title={Titans: Learning to Memorize at Test Time},
author={Ali Behrouz and Peilin Zhong and Vahab Mirrokni},
booktitle={The Thirty-ninth Annual Conference on Neural Information Processing Systems},
year={2025},
url={https://openreview.net/forum?id=8GjSf9Rh7Z}
}

@misc{wang2025testtimeregression,
      title={Test-time regression: a unifying framework for designing sequence models with associative memory}, 
      author={Ke Alexander Wang and Jiaxin Shi and Emily B. Fox},
      year={2025},
      eprint={2501.12352},
      archivePrefix={arXiv},
      primaryClass={cs.LG},
      url={https://arxiv.org/abs/2501.12352}, 
}

@inproceedings{
ma2025tabdptscalingtabularfoundation,
title={Tab{DPT}: Scaling Tabular Foundation Models on Real Data},
author={Junwei Ma and Valentin Thomas and Rasa Hosseinzadeh and Alex Labach and Jesse C. Cresswell and Keyvan Golestan and Guangwei Yu and Anthony L. Caterini and Maksims Volkovs},
booktitle={The Thirty-ninth Annual Conference on Neural Information Processing Systems},
year={2025},
url={https://openreview.net/forum?id=pIZxEOZCId}
}

@inproceedings{katharopoulos2020transformers,
  title={Transformers are rnns: Fast autoregressive transformers with linear attention},
  author={Katharopoulos, Angelos and Vyas, Apoorv and Pappas, Nikolaos and Fleuret, Fran{\c{c}}ois},
  booktitle={International conference on machine learning},
  pages={5156--5165},
  year={2020},
  organization={PMLR}
}

@inproceedings{gu2024mambalineartimesequencemodeling,
title={Mamba: Linear-Time Sequence Modeling with Selective State Spaces},
author={Albert Gu and Tri Dao},
booktitle={First Conference on Language Modeling},
year={2024},
url={https://openreview.net/forum?id=tEYskw1VY2}
}

@inproceedings{kochstate,
  title={State-Space Models for Tabular Prior-Data Fitted Networks},
  author={Koch, Felix and Wever, Marcel and Raisch, Fabian and Tischler, Benjamin},
  booktitle={1st ICML Workshop on Foundation Models for Structured Data},
  year={2025},
}

@inproceedings{hydra,
title={Hydra: Bidirectional State Space Models Through Generalized Matrix Mixers},
author={Sukjun Hwang and Aakash Lahoti and Ratish Puduppully and Tri Dao and Albert Gu},
booktitle={The Thirty-eighth Annual Conference on Neural Information Processing Systems},
year={2024},
url={https://openreview.net/forum?id=preo49P1VY}
}

@inproceedings{
yang2025gateddeltanetworksimproving,
title={Gated Delta Networks: Improving Mamba2 with Delta Rule},
author={Songlin Yang and Jan Kautz and Ali Hatamizadeh},
booktitle={The Thirteenth International Conference on Learning Representations},
year={2025},
url={https://openreview.net/forum?id=r8H7xhYPwz}
}

@inproceedings{schlag2021lineartransformerssecretlyfast,
  author    = {Imanol Schlag and Kazuki Irie and J{\"{u}}rgen Schmidhuber},
  title     = {Linear Transformers Are Secretly Fast Weight Programmers},
  booktitle = {Proceedings of the 38th International Conference on Machine Learning (ICML 2021)},
  series    = {Proceedings of Machine Learning Research},
  pages     = {9355--9366},
  publisher = {{PMLR}},
  year      = {2021},
  url       = {https://proceedings.mlr.press/v139/schlag21a.html}
}

@inproceedings{
baur2024exploration,
title={Exploration of autoregressive models for in-context learning on tabular data},
author={Stefan K. Baur and Sohyeong Kim},
booktitle={NeurIPS 2024 Third Table Representation Learning Workshop},
year={2024},
url={https://openreview.net/forum?id=4dOJ0PRY7R}
}

@article{ismail2019deep,
  title={Deep learning for time series classification: a review},
  author={Ismail Fawaz, Hassan and Forestier, Germain and Weber, Jonathan and Idoumghar, Lhassane and Muller, Pierre-Alain},
  journal={Data mining and knowledge discovery},
  volume={33},
  number={4},
  pages={917--963},
  year={2019},
  publisher={Springer}
}

@article{benavoli2016should,
  title={Should we really use post-hoc tests based on mean-ranks?},
  author={Benavoli, Alessio and Corani, Giorgio and Mangili, Francesca},
  journal={The Journal of Machine Learning Research},
  volume={17},
  number={1},
  pages={152--161},
  year={2016},
  publisher={JMLR. org}
}

@article{wilcoxon1945individual,
  title={Individual comparisons by ranking methods},
  author={Wilcoxon, Frank},
  journal={Biometrics bulletin},
  volume={1},
  number={6},
  pages={80--83},
  year={1945},
  publisher={JSTOR}
}

@article{garcia2008extension,
  title={An Extension on" Statistical Comparisons of Classifiers over Multiple Data Sets" for all Pairwise Comparisons.},
  author={Garcia, Salvador and Herrera, Francisco},
  journal={Journal of machine learning research},
  volume={9},
  number={12},
  year={2008}
}

@article{holm1979simple,
  title={A simple sequentially rejective multiple test procedure},
  author={Holm, Sture},
  journal={Scandinavian journal of statistics},
  pages={65--70},
  year={1979},
  publisher={JSTOR}
}

@article{song2026feat,
  title={FEAT: A Linear-Complexity Foundation Model for Extremely Large Structured Data},
  author={Song, Zhenghang and Qian, Tang and Chen, Lu and Li, Yushuai and Hu, Zhengke and Fang, Bingbing and Song, Yumeng and Zhao, Junbo and Zhang, Sheng and Li, Tianyi},
  journal={arXiv preprint arXiv:2603.16513},
  year={2026}
}

@inproceedings{yang2023gated,
author = {Yang, Songlin and Wang, Bailin and Shen, Yikang and Panda, Rameswar and Kim, Yoon},
title = {Gated linear attention transformers with hardware-efficient training},
year = {2024},
publisher = {JMLR.org},
booktitle = {Proceedings of the 41st International Conference on Machine Learning},
articleno = {2333},
numpages = {23},
location = {Vienna, Austria},
series = {ICML'24}
}

@article{yang2024parallelizing,
  title={Parallelizing linear transformers with the delta rule over sequence length},
  author={Yang, Songlin and Wang, Bailin and Zhang, Yu and Shen, Yikang and Kim, Yoon},
  journal={Advances in neural information processing systems},
  volume={37},
  pages={115491--115522},
  year={2024}
}

@inproceedings{dao2024transformers,
author = {Dao, Tri and Gu, Albert},
title = {Transformers are SSMs: generalized models and efficient algorithms through structured state space duality},
year = {2024},
publisher = {JMLR.org},
booktitle = {Proceedings of the 41st International Conference on Machine Learning},
articleno = {399},
numpages = {31},
location = {Vienna, Austria},
series = {ICML'24}
}

@inproceedings{
bischl2017openml,
title={Open{ML} Benchmarking Suites},
author={Bernd Bischl and Giuseppe Casalicchio and Matthias Feurer and Pieter Gijsbers and Frank Hutter and Michel Lang and Rafael Gomes Mantovani and Jan N. van Rijn and Joaquin Vanschoren},
booktitle={Thirty-fifth Conference on Neural Information Processing Systems Datasets and Benchmarks Track (Round 2)},
year={2021},
url={https://openreview.net/forum?id=OCrD8ycKjG}
}

@inproceedings{
erickson2026tabarena,
title={TabArena: A Living Benchmark for Machine Learning on Tabular Data},
author={Nick Erickson and Lennart Purucker and Andrej Tschalzev and David Holzm{\"u}ller and Prateek Mutalik Desai and David Salinas and Frank Hutter},
booktitle={The Thirty-ninth Annual Conference on Neural Information Processing Systems Datasets and Benchmarks Track},
year={2026},
url={https://openreview.net/forum?id=jZqCqpCLdU}
}

@inproceedings{
buitrago2025understanding,
title={Understanding and Improving Length Generalization in Recurrent Models},
author={Ricardo Buitrago and Albert Gu},
booktitle={Forty-second International Conference on Machine Learning},
year={2025},
url={https://openreview.net/forum?id=2OEb20dy7B}
}

@article{moroshan2025tempopfn,
  title={TempoPFN: Synthetic Pre-training of Linear RNNs for Zero-shot Time Series Forecasting},
  author={Moroshan, Vladyslav and Siems, Julien and Zela, Arber and Carstensen, Timur and Hutter, Frank},
  journal={arXiv preprint arXiv:2510.25502},
  year={2025}
}

@inproceedings{
trockman2025mimetic,
title={Mimetic Initialization Helps State Space Models Learn to Recall},
author={Asher Trockman and Hrayr Harutyunyan and J Zico Kolter and Sanjiv Kumar and Srinadh Bhojanapalli},
booktitle={Workshop on Neural Network Weights as a New Data Modality},
year={2025},
url={https://openreview.net/forum?id=nerNr9fjfD}
}

@inproceedings{
ding2023causallm,
title={Causal{LM} is not optimal for in-context learning},
author={Nan Ding and Tomer Levinboim and Jialin Wu and Sebastian Goodman and Radu Soricut},
booktitle={The Twelfth International Conference on Learning Representations},
year={2024},
url={https://openreview.net/forum?id=guRNebwZBb}
}

@article{vaswani2017attention,
  title={Attention is all you need},
  author={Vaswani, Ashish and Shazeer, Noam and Parmar, Niki and Uszkoreit, Jakob and Jones, Llion and Gomez, Aidan N and Kaiser, {\L}ukasz and Polosukhin, Illia},
  journal={Advances in neural information processing systems},
  volume={30},
  year={2017}
}

@article{grinsztajn2022tree,
  title={Why do tree-based models still outperform deep learning on typical tabular data?},
  author={Grinsztajn, L{\'e}o and Oyallon, Edouard and Varoquaux, Ga{\"e}l},
  journal={Advances in neural information processing systems},
  volume={35},
  pages={507--520},
  year={2022}
}

@article{Vanschoren_2014,
   title={OpenML: networked science in machine learning},
   volume={15},
   ISSN={1931-0153},
   url={http://dx.doi.org/10.1145/2641190.2641198},
   DOI={10.1145/2641190.2641198},
   number={2},
   journal={ACM SIGKDD Explorations Newsletter},
   publisher={Association for Computing Machinery (ACM)},
   author={Vanschoren, Joaquin and van Rijn, Jan N. and Bischl, Bernd and Torgo, Luis},
   year={2014},
   month=jun, pages={49-60} }

@misc{yang2024fla,
  title  = {FLA: A Triton-Based Library for Hardware-Efficient Implementations of Linear Attention Mechanism},
  author = {Yang, Songlin and Zhang, Yu},
  url    = {https://github.com/fla-org/flash-linear-attention},
  month  = jan,
  year   = {2024}
}

@article{grinsztajn2025tabpfn,
  title={Tabpfn-2.5: Advancing the state of the art in tabular foundation models},
  author={Grinsztajn, L{\'e}o and Fl{\"o}ge, Klemens and Key, Oscar and Birkel, Felix and Jund, Philipp and Roof, Brendan and J{\"a}ger, Benjamin and Safaric, Dominik and Alessi, Simone and Hayler, Adrian and others},
  journal={arXiv preprint arXiv:2511.08667},
  year={2025}
}

@article{prokhorenkova2018catboost,
  title={CatBoost: unbiased boosting with categorical features},
  author={Prokhorenkova, Liudmila and Gusev, Gleb and Vorobev, Aleksandr and Dorogush, Anna Veronika and Gulin, Andrey},
  journal={Advances in neural information processing systems},
  volume={31},
  year={2018}
}

@inproceedings{tumma2026preconditioned,
title={Preconditioned DeltaNet: Curvature-aware Sequence Modeling for Linear Recurrences},
author={Neehal Tumma and Noel Loo and Daniela Rus},
booktitle={Forty-third International Conference on Machine Learning},
year={2026},
url={https://openreview.net/forum?id=UC6YiTOeKb}
}

@article{robbins1951stochastic,
  title={A stochastic approximation method},
  author={Robbins, Herbert and Monro, Sutton},
  journal={The annals of mathematical statistics},
  pages={400--407},
  year={1951},
  publisher={JSTOR}
}

@inproceedings{
afzal2026linear,
title={Linear Attention for Efficient Bidirectional Sequence Modeling},
author={Arshia Afzal and Elias Abad Rocamora and Leyla Naz Candogan and Pol Puigdemont and Francesco Tonin and Yongtao Wu and Mahsa Shoaran and Volkan Cevher},
booktitle={The Thirty-ninth Annual Conference on Neural Information Processing Systems},
year={2025},
url={https://openreview.net/forum?id=Ar62cqTduE}
}

@inproceedings{
lenz2025jamba,
title={Jamba: Hybrid Transformer-Mamba Language Models},
author={Barak Lenz and Opher Lieber and Alan Arazi and Amir Bergman and Avshalom Manevich and Barak Peleg and Ben Aviram and Chen Almagor and Clara Fridman and Dan Padnos and Daniel Gissin and Daniel Jannai and Dor Muhlgay and Dor Zimberg and Edden M. Gerber and Elad Dolev and Eran Krakovsky and Erez Safahi and Erez Schwartz and Gal Cohen and Gal Shachaf and Haim Rozenblum and Hofit Bata and Ido Blass and Inbal Magar and Itay Dalmedigos and Jhonathan Osin and Julie Fadlon and Maria Rozman and Matan Danos and Michael Gokhman and Mor Zusman and Naama Gidron and Nir Ratner and Noam Gat and Noam Rozen and Oded Fried and Ohad Leshno and Omer Antverg and Omri Abend and Or Dagan and Orit Cohavi and Raz Alon and Ro'i Belson and Roi Cohen and Rom Gilad and Roman Glozman and Shahar Lev and Shai Shalev-Shwartz and Shaked Haim Meirom and Tal Delbari and Tal Ness and Tomer Asida and Tom Ben Gal and Tom Braude and Uriya Pumerantz and Josh Cohen and Yonatan Belinkov and Yuval Globerson and Yuval Peleg Levy and Yoav Shoham},
booktitle={The Thirteenth International Conference on Learning Representations},
year={2025},
url={https://openreview.net/forum?id=JFPaD7lpBD}
}

\newpage

\appendix

\part{Appendices}%
\etocsettocstyle{}{}%
{\hypersetup{linkcolor=black}\localtableofcontents}%
\newpage

\section{Computational Resources}
\label{app:computational_resources}

Experiments were performed primarily on a cluster node with 8 NVIDIA RTX 2080 Ti GPUs, each with 11 GB of VRAM, and an Intel Xeon Gold 5120 processor. For pretraining with longer context sizes, where GPU memory was a constraint, we used a second cluster node with 8 NVIDIA RTX A5000 GPUs, each with 24 GB of VRAM, and an AMD EPYC 7742 64-Core Processor.

Pretraining each model took approximately one to two days on a single GPU, depending on the underlying architecture. Length-generalisation evaluations additionally required several hours per model. In total for the experiments in the paper, we trained the five linear/recurrent model families across up to four different training setups, as well as causal and non-causal softmax-attention Transformer variants, the mitigation strategies for DeltaNet and GLA, the stabilised DeltaNet variants with final-state readout and write-rate decay, and the layer-local ridge state model. Across all final experiments, this corresponds to approximately 100-125 GPU-days for pretraining and evaluation.

Additional compute was used for preliminary experiments, implementation development, and selecting fair parameter configurations for each architecture.

\section{Training Details} \label{app:training_details}

\subsection{Linear-Time Architecture Details}
\label{app:model_details}

Table~\ref{tab:recurrence_relation_comparison} summarises the  update and readout rules for the evaluated model families.

\begin{table}[!h]
    \centering
    \caption{Unified recurrent-memory view of the linear-time sequence mixers considered in this work. The table is adapted from \citet{yang2024parallelizing}. $\bm{S}_t$ denotes the recurrent memory state, and $\bm{o}_t$ is the unnormalised memory readout. We omit normalisation terms, output projections, and implementation-specific details for clarity.}
    \label{tab:recurrence_relation_comparison}
    \vspace{0.5em}
    \begin{tabular}{lll}
    \toprule
    Model & Recurrence & Readout \\
    \midrule
    Linear Attention
    & $\bm{S}_t = \bm{S}_{t-1} + \bm{v}_t \bm{k}_t^{\top}$
    & $\bm{o}_t = \bm{S}_t \bm{q}_t$ \\
    \quad + Kernel
    & $\bm{S}_t = \bm{S}_{t-1} + \bm{v}_t \phi(\bm{k}_t)^{\top}$
    & $\bm{o}_t = \bm{S}_t \phi(\bm{q}_t)$ \\
    Linear Attn. Gated
    & $\bm{S}_t = \bm{S}_{t-1}\mathrm{Diag}(\alpha_t)
       + \bm{v}_t \bm{k}_t^{\top}$
    & $\bm{o}_t = \bm{S}_t \bm{q}_t$ \\
    Delta Rule
    & $\bm{S}_t = \bm{S}_{t-1}\big(\mathbf{I} - \beta_t \bm{k}_t \bm{k}_t^{\top}\big)
       + \beta_t \bm{v}_t \bm{k}_t^{\top}$
    & $\bm{o}_t = \bm{S}_t \bm{q}_t$ \\
    Delta Rule Gated
    & $\bm{S}_t =
        \bm{S}_{t-1}\big(\alpha_t (\mathbf{I} - \beta_t \bm{k}_t \bm{k}_t^{\top})\big)
       + \beta_t \bm{v}_t \bm{k}_t^{\top}$
    & $\bm{o}_t = \bm{S}_t \bm{q}_t$ \\
    Mamba-2
    & $\bm{S}_t = \gamma_t \bm{S}_{t-1}
       + \bm{v}_t \bm{k}_t^{\top}$ 
    & $\bm{o}_t = \bm{S}_t \bm{q}_t$ \\
    \bottomrule
    \end{tabular}
\end{table}

The size-matched models in Tables~\ref{tab:openml_tabarena_summary} and~\ref{tab:model_comparison_real_world} use embedding dimension $d=320$, 4 heads where applicable, and the feed-forward/intermediate dimension is set to $640=2d$. We set the number of layers separately for each architecture to match all models to approximately 12.5M parameters, resulting in 12 layers for most of the backbones. Table~\ref{tab:matched_parameter_counts} shows the exact parameter counts for the models in these tables.

\begin{table}[!ht]
\centering
\small
\caption{Parameter counts for the models in the controlled comparisons of Tables~\ref{tab:openml_tabarena_summary} and~\ref{tab:model_comparison_real_world}. Counts include the encoders, backbone, and prediction head. Shared weights are counted once. Delta-FS, Delta + Decay, and Delta-FS + Decay have the same parameter count as Delta.}
\label{tab:matched_parameter_counts}
\vspace{0.5em}
\begin{tabular}{lr}
\toprule
Model & Parameters \\
\midrule
Softmax Non-Causal & 12{,}507{,}210 \\
Delta & 12{,}489{,}162 \\
Delta Gated & 12{,}504{,}426 \\
Linear Non-Causal & 12{,}473{,}482 \\
Linear Gated & 12{,}567{,}882 \\
Linear Causal & 12{,}473{,}802 \\
Mamba-2 & 12{,}492{,}582 \\
Layer-Local Ridge & 12{,}473{,}802 \\
Delta + Decay (Int-MT) & 12{,}535{,}242 \\
\bottomrule
\end{tabular}
\end{table}

\subsection{Pretraining and Hyperparameter Details}
\label{app:Pre-training_and_Hyperparameter_details}

Overall, we use a near-identical setup to TabPFN v1, but with different hyperparameters. All models are pretrained on synthetic tasks sampled on the fly from the TabPFN v1 prior. For each minibatch, we sample fresh synthetic datasets, including the number of features, the context/query split, and the number of label classes. As the prior samples new datasets for each minibatch, we used the pretraining loss and a validation loss on additional synthetic datasets for general and backbone-specific hyperparameter selection. The additional synthetic validation datasets are needed for model development, as different training strategies, as described in Section~\ref{subsec:Training_strategies}, provide different access to information, which can change pretraining loss.

Unless stated otherwise, all models use the shared pretraining config described in Table~\ref{tab:pretraining_hyperparameters}. In total, each model sees $4000$ minibatches/epoch $\times$ $200$ epochs $\times$ $8$ datasets/minibatch $=6.4$M synthetic datasets. The loss is always computed on query rows.

\begin{table}[!h]
\centering
\small
\caption{Shared pretraining hyperparameters for the size-matched real-world comparisons.}
\label{tab:pretraining_hyperparameters}
\vspace{0.5em}
\begin{tabular}{ll}
\toprule
Hyperparameter & Value \\
\midrule
Synthetic prior & TabPFN v1 prior \\
Rows per dataset & 1000 \\
Context/query split & uniformly sampled, min. 64 context rows \\
Input features & uniformly sampled from 2-20 \\
Classes & up to 10 \\
Training budget & 200 epochs, 4000 minibatches/epoch \\
Batching & 8 datasets/minibatch, 2-step grad. accumulation \\
Effective batch size & 16 synthetic datasets \\
Embedding/hidden size & 320 \\
Attention heads & 4 \\
FFN/intermediate size & 640 \\
Optimiser & AdamW, lr $3\cdot 10^{-5}$, weight decay $0.01$ \\
Schedule & cosine decay, 10 warmup epochs \\
Loss & cross-entropy on query rows \\
\bottomrule
\end{tabular}
\end{table}

Short-convolution modules are disabled (\texttt{use\_short\_conv=False}), except for DeltaNet Interleaved Multi-Target (Int-MT), which uses convolution width 4. In our exploratory evaluations, short convolutions benefited DeltaNet in the interleaved setting, while reducing performance or providing no material improvement for the other evaluated configurations.

\subsection{External Baselines}
\label{app:Baselines_used_in_Evaluation}

To provide reference performance on our real-world benchmarks, we include TabPFN v2.5 \citep{grinsztajn2025tabpfn, hollmann2025tabpfn}, TabICLv2 \citep{jingangtabicl, qu2026tabiclv2}, TabFlex \citep{zeng2025tabflex}, and CatBoost \citep{prokhorenkova2018catboost}.

TabPFN v2.5 and TabICLv2 represent the state of the art in tabular foundation models according to TabArena \citep{erickson2026tabarena}. TabFlex is the most closely aligned baseline, as it uses a non-causal linear attention backbone and is also trained on the TabPFN v1 prior \citep{hollmann2023tabpfn}. Importantly, these baselines are not controlled for training budget or model size: we evaluate the models released by their respective authors. The foundation model baselines have comparable or larger parameter counts than our 12.5M-parameter models, and were pretrained on substantially more data, including larger datasets. Our models were pretrained on 6.4B rows of data, while we estimate the pretraining data exposure for these baseline models at 40B-100B+ rows. These differences, together with the more advanced priors used by TabICLv2 and TabPFN v2.5, help explain the performance gap in Table~\ref{tab:openml_tabarena_summary} between our controlled models and the external baselines.

CatBoost is included as a strong gradient-boosted decision-tree baseline. As shown in the TabArena evaluation by \citet{erickson2026tabarena}, CatBoost performs very well out of the box without hyperparameter tuning, supporting its use as a default configuration baseline.

\subsection{Hidden-State Oracle Details}
\label{app:Hidden_state_oracle_details}

For each evaluation dataset, the hidden-state oracle proceeds as follows. We first run a selected linear-time model (in this case, the pretrained Comb-ST DeltaNet) on the context samples to obtain an initial hidden state $\bm{S}$ for each layer. We then freeze all model parameters and optimise only these hidden states using multi-epoch mini-batch gradient descent on the end-to-end test-time regression objective over the context set
\begin{equation}
    \mathcal{L}(\bm{S}) = 
    \mathbb{E}_{(x_i, y_i) \sim \mathcal{C}} 
    \mathrm{CE}(f_{\theta}(\bm{S}, x_i), y_i)
    .
\end{equation}
Here, $f_{\theta}(\bm{S}, x_i)$ denotes the prediction of the frozen model for a context row $x_i$ that is processed as a query. Each layer reads from its own state in $\bm{S}$, but since the loss is backpropagated through the whole network, the states of all layers are optimised jointly. This procedure keeps the model architecture, projections, and readout fixed, modifying only the per-layer fast-weight states $\bm{S}$.

For hyperparameter tuning and early stopping, we split the context set into training and validation subsets, holding out 10\% of the context rows. Early stopping is necessary because the optimised hidden states can overfit the context set, even at a sequence length of 128k. This further supports the conclusion that fixed-size hidden-state capacity is not the primary bottleneck.

\subsection{Layer-Local Ridge State Details}
\label{app:layer_local_ridge_details}

The layer-local ridge state model replaces the recurrent state update with the closed-form solution of a length-scaled ridge objective. This deviates from the hidden-state oracle by not optimising the hidden states jointly across all layers on the end-to-end objective but by computing the state of each layer independently from the layer-local objective of Equation~\ref{eq:ttr_objective} and using its closed-form solution. Additionally, unlike the hidden-state oracle, which uses frozen model parameters, we pretrain the layer-local ridge state model from scratch and match the pretraining budget (results in Appendix~\ref{app:layer_local_ridge_comparison}). Let $\bm{K}\in\mathbb{R}^{n\times d_{qk}}$ and $\bm{V}\in\mathbb{R}^{n\times d_v}$ collect the keys and values as rows. The Moore-Penrose solution $\bm{S}_n^\star = \bm{V}^\top (\bm{K}^\dagger)^\top$ to the layer-local least-squares problem corresponds to the least-squares memory of \citet{wang2025testtimeregression}. In our experiments and in line with \citet{tumma2026preconditioned}, learned key matrices can become ill-conditioned, making the ridgeless solution unstable and increasing both pretraining loss and long-context degradation. We therefore use a length-scaled ridge objective with closed-form solution
\begin{equation*}
    \bm{S}_n^{\mathrm{ridge}}
    = \operatorname*{argmin}_{\bm{S} \in \mathbb{R}^{d_v \times d_{qk}}} \frac{1}{n} \sum_{t=1}^n
        \left\|
            \bm{v}_t - \bm{S}\bm{k}_t
        \right\|_2^2 + \lambda \left\| \bm{S} \right\|_F^2
    =
    \bm{V}^\top \bm{K}
    \left(\bm{K}^\top \bm{K}+n\lambda \bm{I}\right)^{-1}.
\end{equation*}
This state accounts for the covariance structure of key representations, remains a permutation-invariant function of the context, and scales the regularisation strength with the context length $n$.

\section{Evaluation Details} \label{app:evaluation_details}
\subsection{Metrics Used in Evaluation}
\label{app:Metrics_used_in_evaluation}

In the main evaluation, we report raw ROC-AUC and accuracy, averaged over splits and datasets. Details on significance testing are provided in Appendix~\ref{app:significance_testing_details}. For the length generalisation experiments, we sample 500 synthetic datasets of 128k rows from the pretraining prior and truncate them to context prefixes of increasing length $n_c$, holding the query set fixed at $n_q=100$ samples. We report normalised ROC-AUC and accuracy. Since the synthetic datasets vary substantially in difficulty, normalisation makes confidence intervals more comparable when aggregating across datasets.

The shaded regions in the plots show 95\% confidence intervals. Normalisation constants are computed per dataset using a fixed reference set consisting of the controlled model families from Table~\ref{tab:openml_tabarena_summary}, including the four-head softmax reference, and the hidden-state oracle, rather than only the models shown in each figure. For normalisation, all evaluated context lengths of the same synthetic dataset share the same normalisation constants. This makes scores comparable across the different length-generalisation figures.

Specifically, for each dataset $d$, metric $m$, and model/context-length configuration $r \in \mathcal{R}$, we normalise raw scores using the minimum over reference configurations at context lengths up to 1k ($\mathcal{R}_{\leq 1\mathrm{k}}$) and the maximum over all reference configurations:
\begin{equation}
    \hat{s}_{d,r}^{(m)}
    =
    \frac{s_{d,r}^{(m)} - \min_{r' \in \mathcal{R}_{\leq 1\mathrm{k}}} s_{d,r'}^{(m)}}
    {\max_{r' \in \mathcal{R}} s_{d,r'}^{(m)} - \min_{r' \in \mathcal{R}_{\leq 1\mathrm{k}}} s_{d,r'}^{(m)}} .
\end{equation}

\subsection{Significance Testing}
\label{app:significance_testing_details}

Our significance testing follows the pairwise comparison protocol used by \citet{ismail2019deep}, which, in turn, is based on the recommendation of \citet{benavoli2016should} to use tests that perform pairwise comparisons, such as the Wilcoxon signed-rank test. Accordingly, for each benchmark and metric, we perform paired two-sided Wilcoxon signed-rank tests \citep{wilcoxon1945individual} and additionally apply Holm's alpha correction at $\alpha = 5\%$ \citep{holm1979simple, garcia2008extension}.

\subsubsection{Significance Testing for Training Strategies}

Let $x_{f,d,s,r}^{(m)}$ denote the value of metric $m$ for model family $f$, dataset $d$, training setup $s$, and cross-validation split $r$.

For this analysis, we include only model families for which all four training setups are available. In our experiments, these are Gated Linear Attention, DeltaNet, Gated DeltaNet, and Mamba-2. We exclude standard Linear Attention because the interleaved-label setup would require custom positional encodings to obtain comparable performance.

\begin{enumerate}
    \item Average over the cross-validation splits:
    \[
        \bar{x}_{f,d,s}^{(m)} = \frac{1}{R}\sum_{r=1}^{R} x_{f,d,s,r}^{(m)}
    \]
    \item For each dataset and setup, we then average over the model families:
    \[
       \tilde{x}_{d,s}^{(m)} = \frac{1}{F}\sum_{f=1}^{F} \bar{x}_{f,d,s}^{(m)}
    \]
    \item For each benchmark and metric, we use Wilcoxon signed-rank tests with Holm correction to compare two setups $s$ and $s'$ pairwise.
\end{enumerate}

\subsubsection{Significance Testing for Sequence Mixers}

For the recurrence formulation comparison, we used the same paired Wilcoxon/Holm procedure described above. The training setup is Combined Single-Target, except for the configurations labelled Int-MT. For each model, dataset, benchmark, and metric, we first average over cross-validation splits. We then compare all pairs of included models using the resulting dataset-level paired samples. Holm correction is applied separately for each benchmark and metric: across the 21 pairs of Table~\ref{tab:openml_tabarena_summary}, the 21 pairs of Table~\ref{tab:model_comparison_real_world}, and all 55 pairs in Tables~\ref{tab:extended_pairwise_openml} and~\ref{tab:extended_pairwise_tabarena}. Because each table corrects within its own set of models, the same model pair can receive different significance decisions across tables. Shared bars indicate non-significant differences, not equivalence.

\newpage
\section{Additional Results}

\subsection{Real-World Length Generalisation on TabArena} \label{app:realworld_length_generalisation}

\begin{figure}[!h]
    \centering
    \includegraphics[width=\linewidth, trim={0pt, 0pt, 0pt, 25pt}, clip]{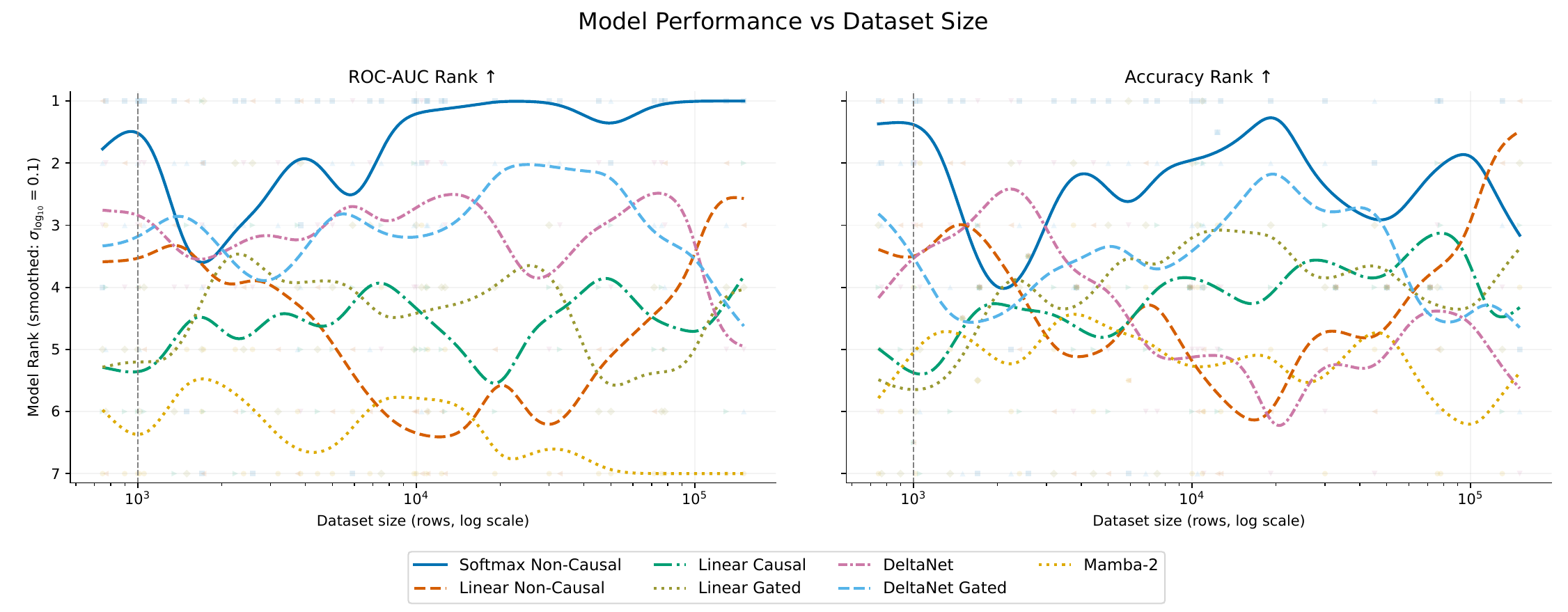}
    \caption{Model ranks as a function of dataset size in TabArena. The ranks for each model are smoothed in $\mathrm{log10}$-space with a Gaussian kernel of $\sigma_{\mathrm{log10}}=0.1$. At moderate sizes, DeltaNet variants are competitive with softmax attention. With increasing dataset size $n_c$, non-causal softmax and linear attention outperform the causal models.}
    \label{fig:real_world_seqlen}
\end{figure}

Sequence length generalisation results on the TabArena benchmark across selected models are shown in Figure~\ref{fig:real_world_seqlen}. This visualises the effect of causal degradation in a long-context real-world task.

\subsection{Hidden-State Oracle on Real Data}
\label{app:real_data_oracle}

The hidden-state oracle of Section~\ref{sec:understanding_generalization_failure} is evaluated on datasets sampled from our synthetic prior. We here repeat the analysis on the five largest TabArena classification datasets, using the same procedure as in Appendix~\ref{app:Hidden_state_oracle_details} and stratified 5-fold cross-validation.

Table~\ref{tab:real_data_oracle} reports mean ROC-AUC across folds. The oracle outperforms causal DeltaNet, non-causal linear attention, and non-causal softmax attention on every dataset and on mean performance. The capacity conclusion drawn from the synthetic prior therefore also holds on real-world data: a fixed-size state reaching strong performance exists at these context lengths.

\begin{table}[!h]
\centering
\setlength{\tabcolsep}{5pt}
\caption{Hidden-state oracle comparison on the five largest TabArena classification datasets, reporting ROC-AUC averaged over stratified 5-fold cross-validation splits. Bold marks the best value per dataset and best mean.}
\label{tab:real_data_oracle}
\vspace{0.5em}
\begin{tabular}{lrcccc}
\toprule
Dataset & Rows & DeltaNet & \shortstack{Hidden-State\\Oracle} & \shortstack{Softmax\\Non-Causal} & \shortstack{Linear\\Non-Causal} \\
\midrule
GiveMeSomeCredit      & 150{,}000 & 0.8446 & \textbf{0.8615} & 0.8559 & 0.8546 \\
Airline satisfaction  & 129{,}880 & 0.9314 & \textbf{0.9931} & 0.9765 & 0.9486 \\
SDSS17                & 78{,}053  & 0.9851 & \textbf{0.9938} & 0.9915 & 0.9889 \\
APSFailure            & 76{,}000  & 0.9828 & \textbf{0.9862} & 0.9845 & 0.9816 \\
Diabetes130US         & 71{,}518  & 0.5587 & \textbf{0.6165} & 0.5887 & 0.5418 \\
\midrule
Mean                  &           & 0.8605 & \textbf{0.8902} & 0.8794 & 0.8631 \\
\bottomrule
\end{tabular}
\end{table}

\subsection{Layer-Local Ridge State Comparison}
\label{app:layer_local_ridge_comparison}

\begin{figure}[!h]
    \centering
    \includegraphics[width=.95\textwidth, trim={5pt, 0pt, 7pt 0pt}, clip]{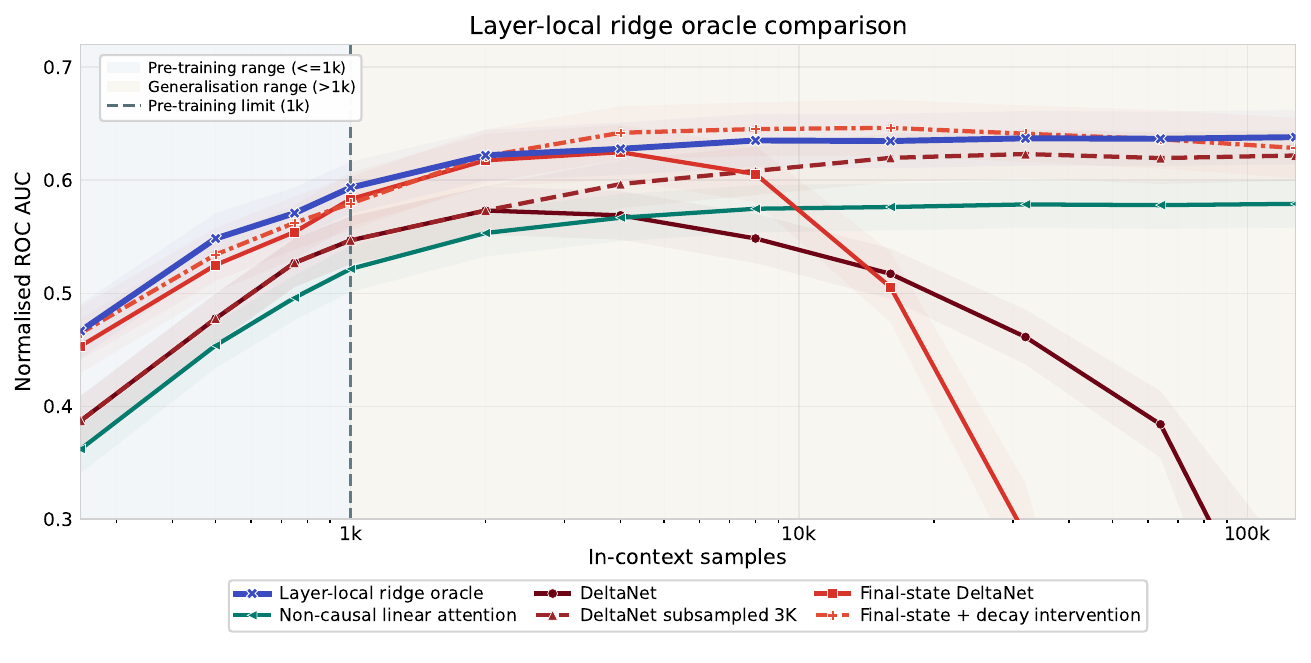}
    \caption{Comparison of layer-local ridge model with DeltaNet variants and non-causal linear attention over increasing sequence lengths. Curves show globally normalised ROC-AUC with 95\% confidence intervals. The layer-local ridge model is pretrained from scratch using the same protocol, but replaces the recurrent state update with a closed-form ridge regression solve.}
\label{fig:ridge_oracle_sequence_length_generalisation}
\end{figure}

Figure~\ref{fig:ridge_oracle_sequence_length_generalisation} shows the synthetic length-generalisation comparison between the layer-local ridge state model (Appendix~\ref{app:layer_local_ridge_details}), the DeltaNet variants, and non-causal linear attention. Pretrained under the same protocol as the recurrent models, the ridge state is length-stable and substantially improves upon non-causal additive linear attention, yet the final-state readout DeltaNet with write-rate decay closely matches and sometimes exceeds it over the evaluated range. If DeltaNet's success were exclusively due to approximating the closed-form solution of the layer-local objective, replacing its update with the ridge solution should be expected to improve performance. This suggests that the layer-local associative-memory objective is not a complete proxy for downstream tabular ICL performance.

The full benchmark comparison, including the layer-local ridge state model, is shown in Tables~\ref{tab:extended_pairwise_openml} and~\ref{tab:extended_pairwise_tabarena}. The ridge state performs competitively but is outperformed by the stabilised DeltaNet, with significant differences on TabArena.

\subsection{Scaling Pretraining Sequence Length}

\begin{figure}[!h]
    \centering
    \includegraphics[width=0.9\textwidth]{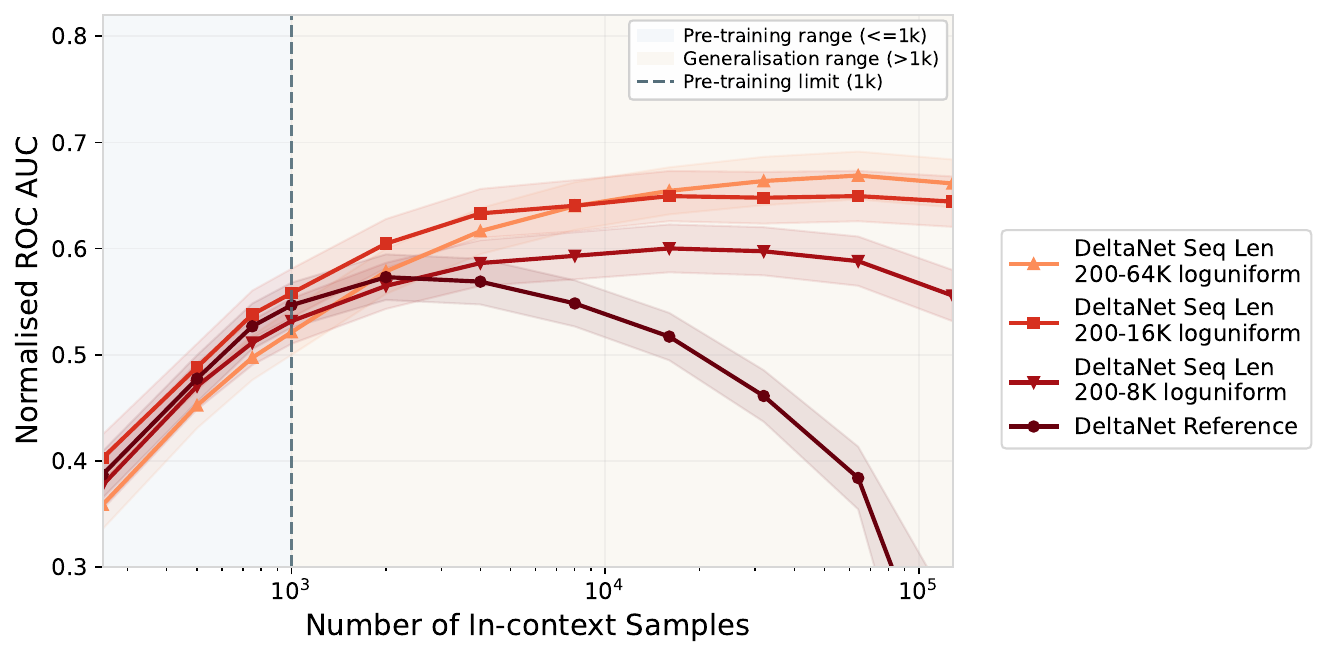}
    \caption{Length generalisation of DeltaNet models pretrained with longer context lengths. The reference model is pretrained with at most 1000 rows, while the other models sample the row number from a log-uniform distribution between 200 and the maximum shown in the legend. Longer-context pretraining improves long-context performance but mainly shifts degradation to larger sequences. The comparison is not compute-matched, since all models see the same number of datasets but a different total number of rows. A compute-matched version is in Section~\ref{sec:understanding_generalization_failure} of the main paper.}
    \label{fig:deltanet_high_seq_len_training_seq_len_comparison}
\end{figure}

Figure~\ref{fig:deltanet_high_seq_len_training_seq_len_comparison} shows length generalisation when pretraining DeltaNet with longer context lengths. For all models except the DeltaNet reference, we sample the number of rows per pretraining dataset from a log-uniform distribution between 200 and a maximum of 8k, 16k, or 64k rows. 

In the figure, we see that with longer-context pretraining, we improve performance at longer evaluation contexts as expected. However, this does not address the underlying issue; it only shifts the degradation to longer sequence lengths. Importantly, these runs are not compute-matched. Although all models are pretrained on the same number of datasets, long-context models see substantially more rows during pretraining. This also explains why their short-context performance does not deteriorate, unlike in the compute-matched setting in Figure~\ref{fig:causal_and_oracle_seq_len_comparison}.

\subsection{Generalisation and Mitigation Strategies for GLA}
\label{app:Generalisation_and_mitigation_strategies_for_gla}

\begin{figure}[!h]
    \centering
    \includegraphics[width=\textwidth]{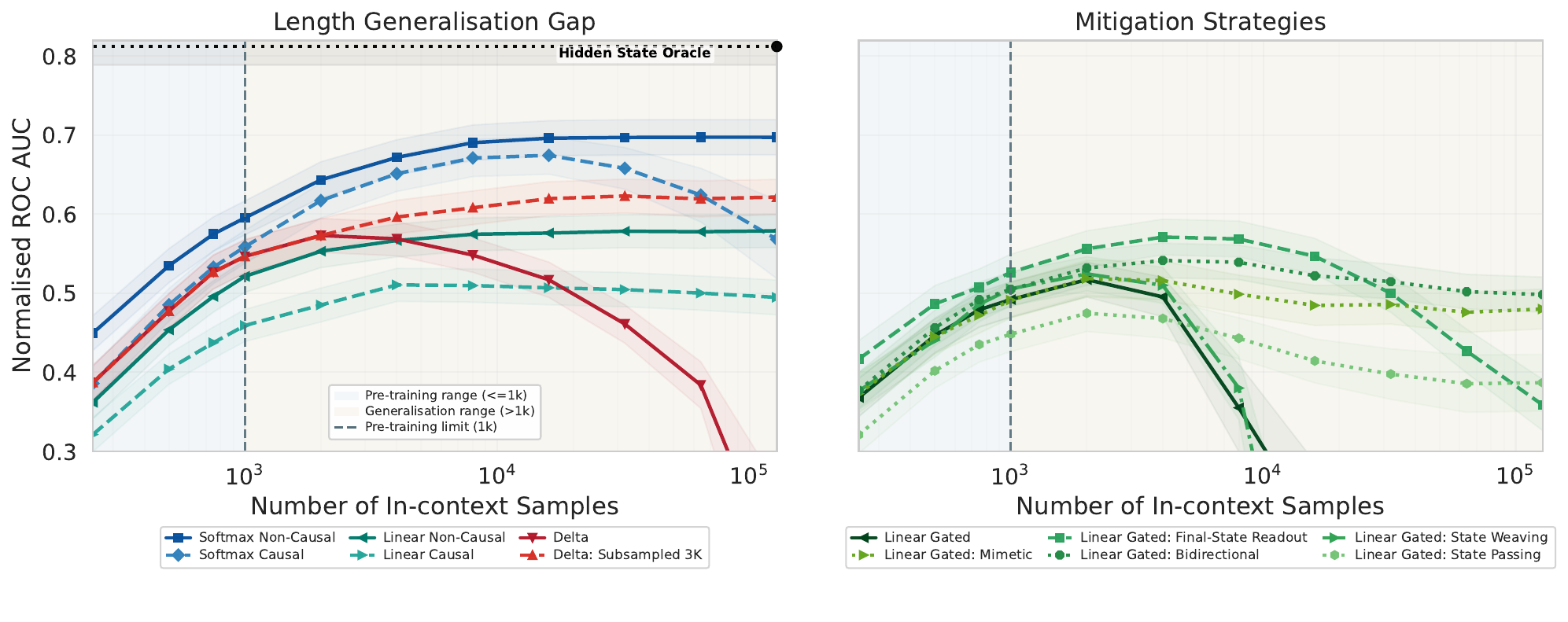}
    \caption{Length generalisation for GLA beyond the 1k pretraining length. We report normalised ROC-AUC averaged over 500 synthetic datasets with 95\% confidence intervals.
    \textbf{(a) Length Generalisation Gap:} Compared to other models, GLA exhibits strong degradation starting at $4\times$ the pretraining context length.
    \textbf{(b) Mitigation Attempts:} State passing and mimetic initialisation mitigate but do not eliminate long-sequence degradation, while bidirectionality is the strongest mitigation for GLA at very long contexts. Final-state readout is strongest up to around $16\times$ the pretraining length but degrades substantially beyond it. State weaving slightly improves performance in the pretraining context range, but shows no significant improvement in addressing degradation.}
    \label{fig:gla_causal_and_oracle_seq_len_comparison}
\end{figure}

Figure~\ref{fig:gla_causal_and_oracle_seq_len_comparison} visualises the same length-generalisation analysis as Section~\ref{subsec:Mitigation_strategies} for gated linear attention (GLA). For GLA, the mitigation strategies are more effective than for DeltaNet, although they do not fully remove the degradation.

\textit{State passing} \citep{buitrago2025understanding}, which exposes the model to longer effective histories by reusing the cached recurrent states from prior batches, improves long-context behaviour. However, it still introduces degradation and reduces the performance within the pretraining range. \textit{State weaving} \citep{moroshan2025tempopfn}, which introduces non-causal information flow by initialising the hidden state of the subsequent layer with the final hidden state of the previous layer, shows minor improvements at pretraining context sizes, but does not reduce length generalisation degradation. \textit{Mimetic initialisation} \citep{trockman2025mimetic}, which initialises gates to be open by default and makes GLA behave closer to ungated additive linear attention, also reduces degradation. \textit{Bidirectionality} and \textit{final-state readout}, which both remove the readout-level causal bottleneck, are the most effective overall. Final-state readout is strongest up to around $16\times$ the pretraining length, after which bidirectionality overtakes it and remains the strongest at very long contexts.

We hypothesise that these improvements are mostly explained by moving GLA closer to causal linear attention through mimetic initialisation, while bidirectionality and final-state readout move its information flow towards a non-causal formulation. Both causal and non-causal linear attention exhibit less degradation than the underlying GLA, consistent with the more stable long-context behaviour of these variants.

\newpage
\subsection{Generalisation Failure Analysis for Linear Attention}
\label{app:Generalisation_Failure_Analysis}

\begin{figure}[!h]
    \centering
    \includegraphics[width=0.8\textwidth, trim={17pt, 0pt, -30pt, 0pt}]{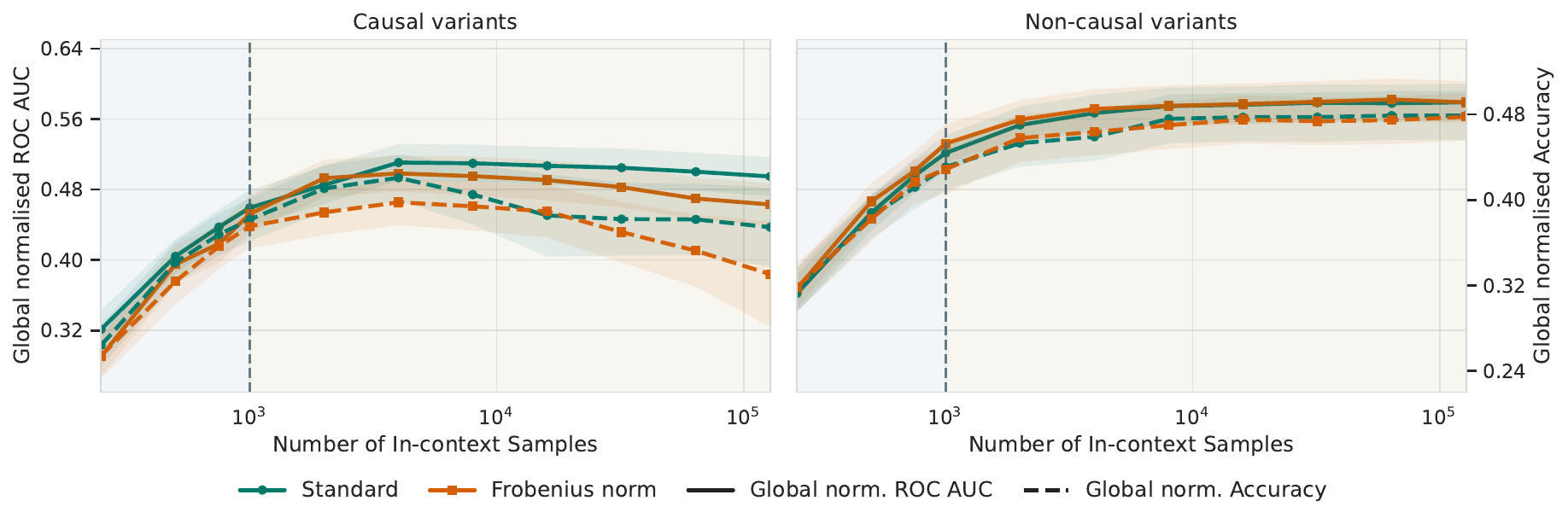}
    \includegraphics[width=0.82\textwidth, trim={0pt, 0pt, -112pt, 0pt}]{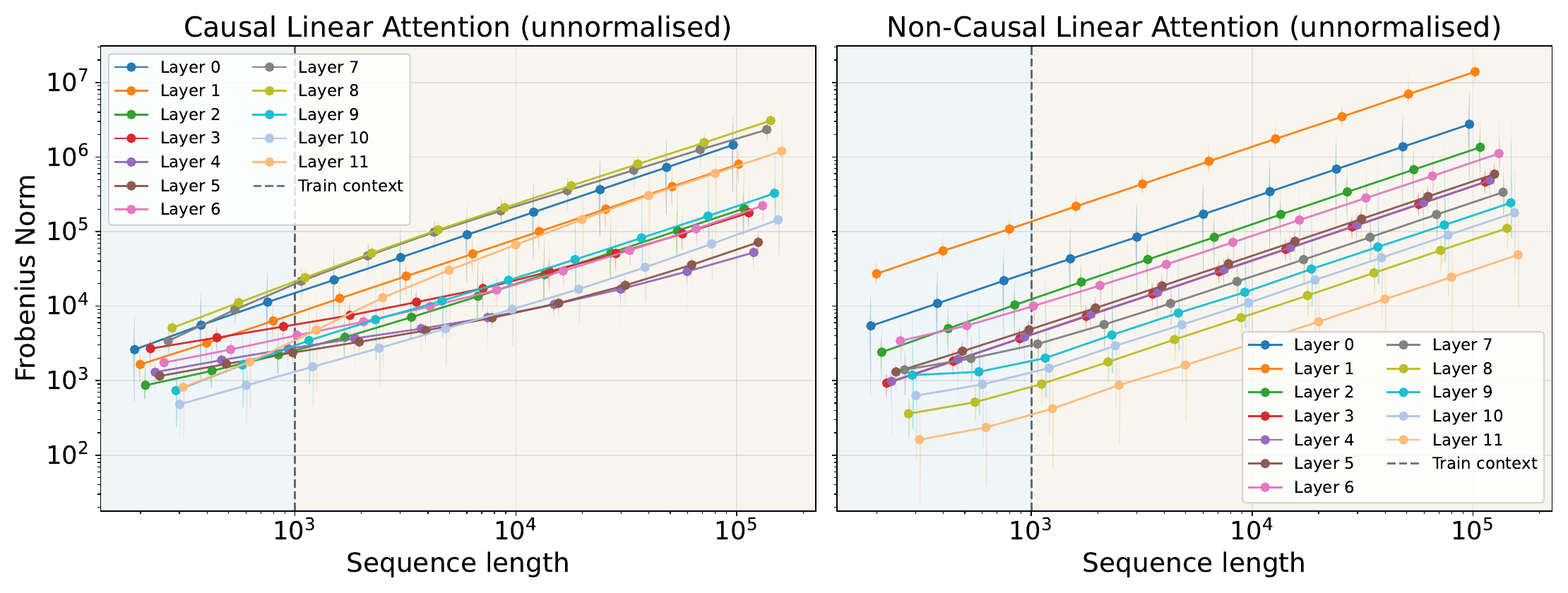}
    \includegraphics[width=0.8\textwidth, trim={11pt, 0pt, -11pt, 0pt}]{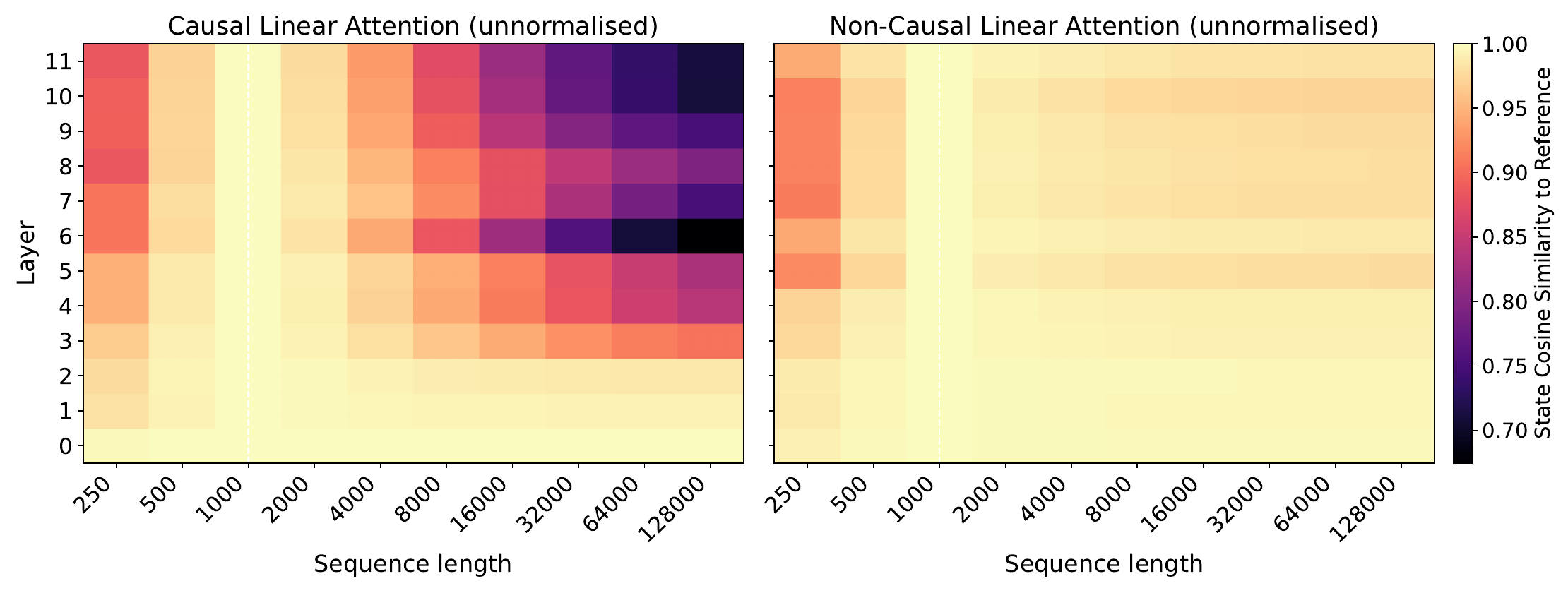}
    \includegraphics[width=0.82\textwidth, trim={0pt, 0pt, -112pt, 0pt}]{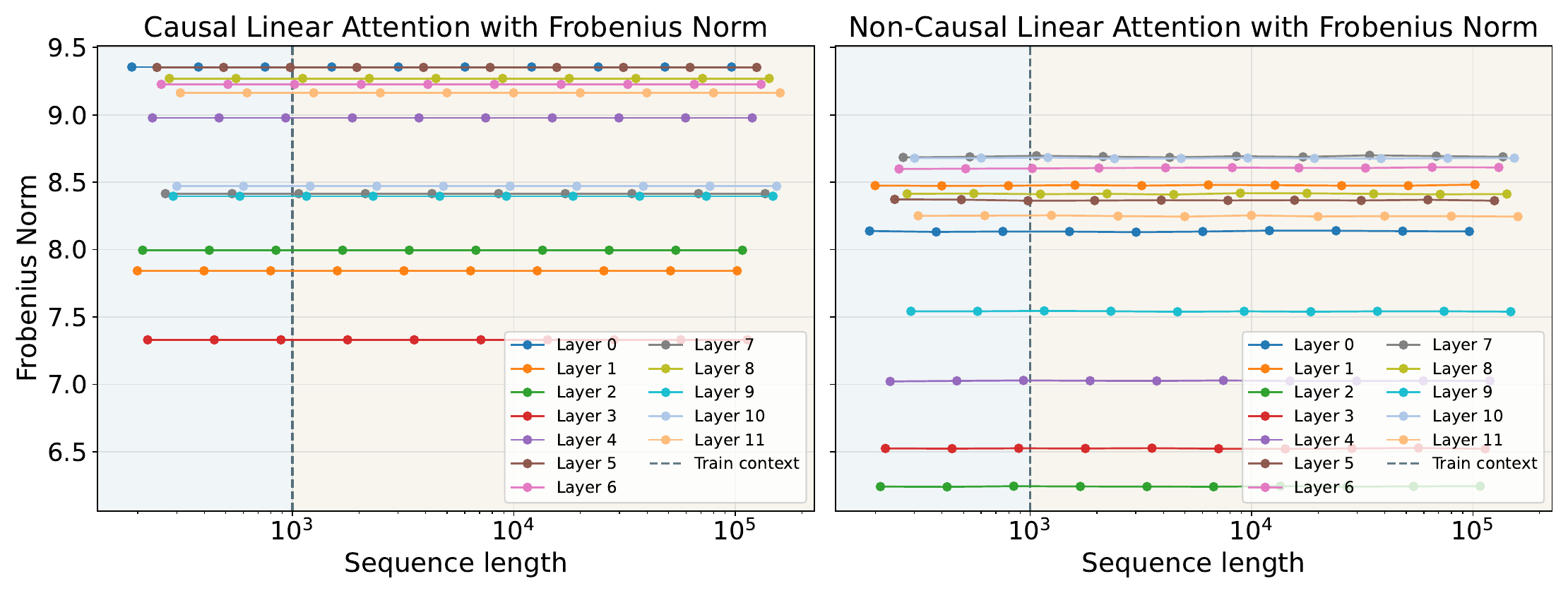}
    \includegraphics[width=0.8\textwidth, trim={11pt, 0pt, -11pt, 0pt}]{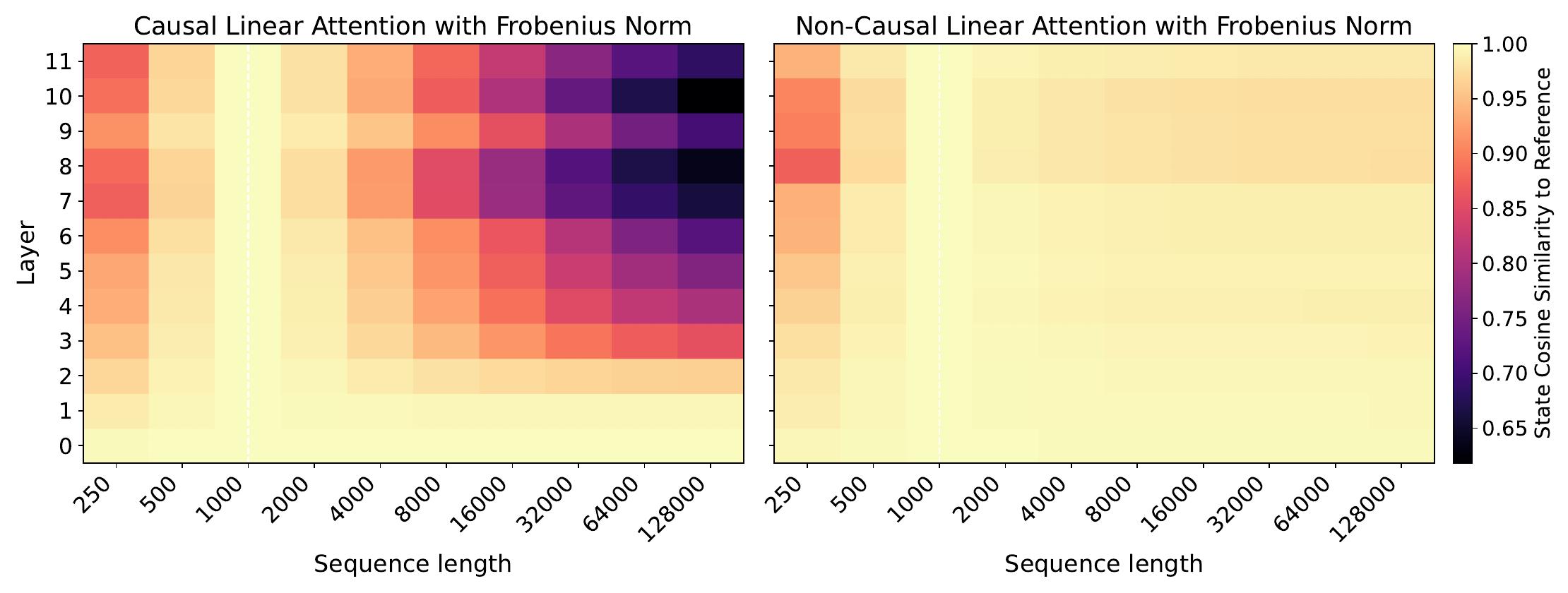}
    \caption{Length generalisation performance of standard and Frobenius-normalised causal and non-causal linear attention variants (top panel), and Frobenius norm evolution for the first head, together with directional drift averaged over heads and repetitions, across sequence lengths, for the unnormalised (second and third panels) and Frobenius-normalised (bottom two panels) models. Constraining the state magnitude does not remove the degradation of the causal model.}
    \label{fig:hidden_state_debug_frobenius_unnormalised}
\end{figure}

When comparing causal linear attention to its non-causal counterpart, we observe that the Frobenius norm of the recurrent key-value state drifts relative to other layers with increasing context length in the causal setup, while this effect is much weaker in the non-causal setup, as shown in Figure~\ref{fig:hidden_state_debug_frobenius_unnormalised}. This behaviour is consistent both when using the K-sum denominator normalisation $\frac{1}{Z_i}$ from Equation~\ref{eq:linear-attention} and when omitting it. We controlled for this magnitude drift by explicitly renormalising the recurrent key-value state. For a linear-attention layer $\ell$ and head $h$, let $S_{\ell,h,t} \in \mathbb{R}^{d_v \times d_{qk}}$ denote the recurrent key-value state after processing context position $t$. At readout time, we use the Frobenius-normalised state:
\begin{equation}
    \widetilde{S}_{\ell,h,t} = S_{\ell,h,t} \cdot \frac{\alpha_{\ell,h}\sqrt{d_v}}
{\max(\|S_{\ell,h,t}\|_F, \varepsilon)}.
\end{equation}

Here $\|S_{\ell,h,t}\|_F = \sqrt{\sum_{i,j} \left(S_{\ell,h,t}\right)_{ij}^2}$ denotes the Frobenius norm, $\varepsilon$ is a small numerical constant, and $\alpha_{\ell,h}$ is a learned per-layer, per-head scale factor initialised to one and parametrised in log-space.

The normalisation preserves the direction of the recurrent state matrix while controlling its magnitude. The readout then uses the normalised state rather than the unnormalised state. We find that this constraint has only a very limited effect on both causal and non-causal linear attention performance (top panel of Figure~\ref{fig:hidden_state_debug_frobenius_unnormalised}).
For the retrained models with the Frobenius-normalised recurrent states, the readout uses a state whose magnitude is constrained, so remaining changes in the effective state are primarily directional. We therefore measure hidden-state drift by comparing the orientation of the recurrent key-value state at each sequence length to the corresponding state at the 1000-row pretraining context length. For each model, repetition, layer, and head, we flatten the effective recurrent state matrix and compute its cosine similarity to the same layer and head at sequence length 1000. The heatmaps in Figure~\ref{fig:hidden_state_debug_frobenius_unnormalised} report the average of cosine similarity over heads and repetitions for each layer and sequence. For causal linear attention the cosine similarity decreases steadily with context length, most strongly in deeper layers, whereas for the non-causal model it stays close to one. This indicates that the Frobenius-normalised causal state drifts away from its short-context configuration as the context grows, whereas the non-causal state remains close to its short-context configuration.

\clearpage

\subsection{DeltaNet Update Dynamics Analysis}
\label{app:deltanet_update_dynamics}

We here provide the full analysis of the DeltaNet update dynamics summarised in Section~\ref{subsec:recurrent_state-estimator}.

\textbf{Linear attention vs DeltaNet state construction.} The additive linear-attention state $\bm{S}_{n} = \sum_{t=1}^{n} \bm{v}_t \bm{k}_t^\top = \bm{V}^\top \bm{K}$ is a commutative set aggregate for fixed layer-local keys and values and does not depend on the arbitrary order of the context rows. While the raw sum grows with $n$, its length-normalised version converges under i.i.d.\ sampling to the expected key-value outer product $\mathbb{E}[\bm{v}\bm{k}^\top]$, so with appropriate state or readout normalisation it behaves like a stable set statistic. It can also be interpreted as a single gradient step from a zero-initialised state on the layer-local memory objective of Equation~\ref{eq:ttr_objective} \citep{wang2025testtimeregression}. The delta-rule state, in contrast, is not an additive set aggregate but an ordered residual update. In the test-time regression view \citep{wang2025testtimeregression}, DeltaNet performs first-order updates on the per-sample squared loss:
\begin{equation*}
    \bm{S}_t = \bm{S}_{t-1}\big(\mathbf{I} - \beta_t \bm{k}_t \bm{k}_t^{\top}\big)
           + \beta_t \bm{v}_t \bm{k}_t^{\top}
    = \bm{S}_{t-1} - \beta_t \nabla_{\bm{S}}\ell_t(\bm{S}_{t-1}), \qquad
    \ell_t(\bm{S}) = \frac{1}{2}
    \left\|
        \bm{S}\bm{k}_t - \bm{v}_t
    \right\|_2^2 .
\end{equation*}
Therefore, final-state readout does not make DeltaNet fully order-independent. For additive linear attention, replacing prefix states with the final state recovers the non-causal set-aggregate structure, whereas for DeltaNet the state $\bm{S}_{n}$ itself remains the result of an ordered online optimisation trajectory.

\textbf{Learned write rates and effective memory.} One central learned component of the delta-rule update is the input-dependent write rate $\beta_t = \sigma(\bm{w}_\beta^\top \bm{h}_t)$. For a fixed unit-norm key direction, the update changes the memory readout from its previous value towards the current value:
\begin{equation*}
    \bm{S}_t \bm{k}_t = (1-\beta_t)\bm{S}_{t-1}\bm{k}_t + \beta_t \bm{v}_t.
\end{equation*}
Thus, a larger $\beta_t$ gives more influence to the current sample and overwrites the previous memory more strongly along the key direction, while a smaller $\beta_t$ preserves more of the previous state. The effective memory horizon of DeltaNet is therefore controlled by the learned sequence of write rates, which can overfit to the pretraining range and be overly aggressive when the same recurrence is unrolled over much larger sequence lengths.

\begin{figure}[!h]
    \makebox[\textwidth][l]{
        \includegraphics[
            width=.955\textwidth,
            trim={3pt 6pt 0pt 0pt},
            clip
        ]{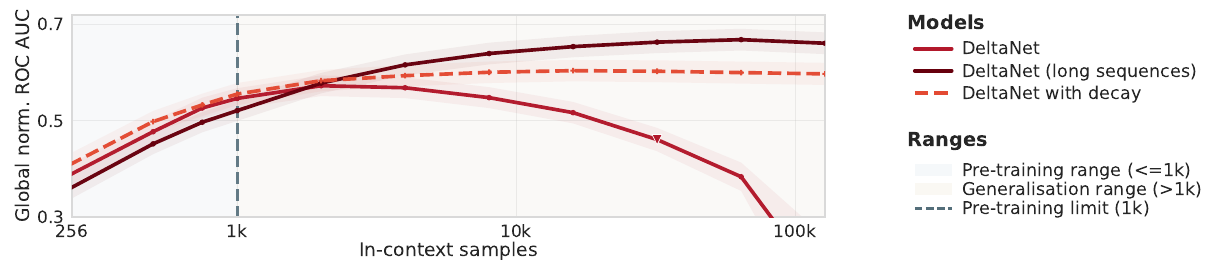}
    }
    \centering
    \includegraphics[width=\textwidth, trim={3pt, 0pt, 0pt 0pt}, clip]{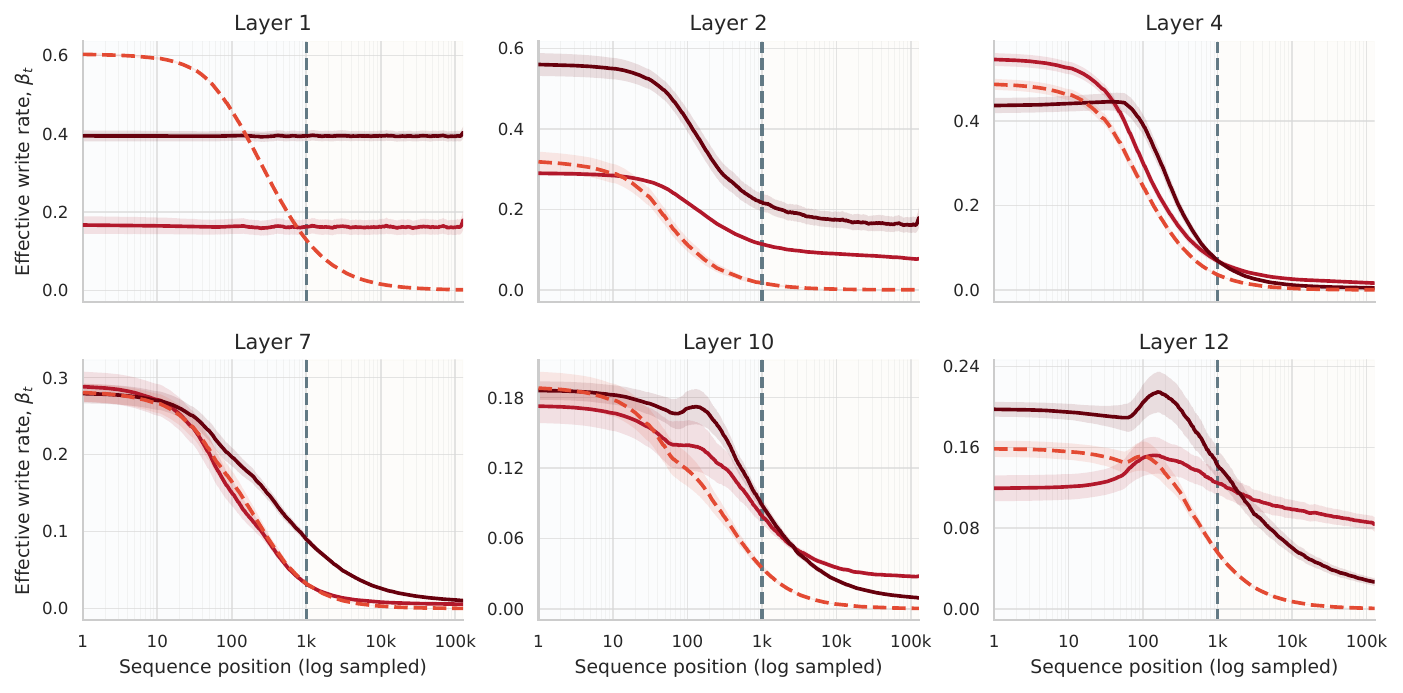}
    \caption{Effective DeltaNet write rate $\beta_t$ across sequence positions and its relation to length generalisation. The top panel shows globally normalised ROC-AUC as the number of in-context samples increases. The lower panels show effective write rates $\beta_t$ across sequence positions for selected layers of DeltaNet, a variant pretrained on longer sequences, and the write-rate decay intervention. For the long-sequence pretraining variants, dataset sizes are sampled log-uniformly between 200 and 64k rows. Write-rate values are averaged across 50 datasets, with means and 95\% confidence intervals visualised. Standard DeltaNet learns a decreasing write-rate profile in deeper layers, where long-sequence pretraining further lowers write rates at later positions. The write-rate decay explicitly decays the effective write rate towards zero across the sequence length in every layer.}
    \label{fig:deltanet_beta_over_seq_len_full}
\end{figure}

Figure~\ref{fig:deltanet_beta_over_seq_len_full} compares the effective write rates of standard DeltaNet, a variant pretrained on sequence lengths of up to 64,000 samples instead of 1,000, and the write-rate decay intervention, extending the layer view of Figure~\ref{fig:deltanet_beta_over_seq_len} in the main paper. The long-context pretrained models are not compute-matched and are used only for analysis. The standard DeltaNet variants learn a decreasing write-rate schedule from early layers onward. Since the models do not use explicit positional encodings, later layers need to infer positional information from the evolving causal representations produced by earlier layers. Long-context pretraining increases this effect and generally yields lower effective write rates at later sequence positions and inside deeper layers. This pattern is consistent with improved long-context performance, suggesting that reduced late-sequence write rates can help preserve earlier information and reduce degradation.

The write-rate decay explicitly decays the effective write rate towards zero across the sequence length in every layer, whereas both learned schedules reduce the write rates with long sequences. Especially in deeper layers, DeltaNet reduces its write rates far less aggressively than the long-sequence version and the decay variant. The top panel shows the resulting length generalisation, where standard DeltaNet collapses beyond the pretraining range while the decay variant remains nearly stable across the full evaluation range.

\begin{figure}[!h]
    \centering
    \includegraphics[width=0.9\textwidth, trim={0pt, 0pt, 0pt 0pt}, clip]{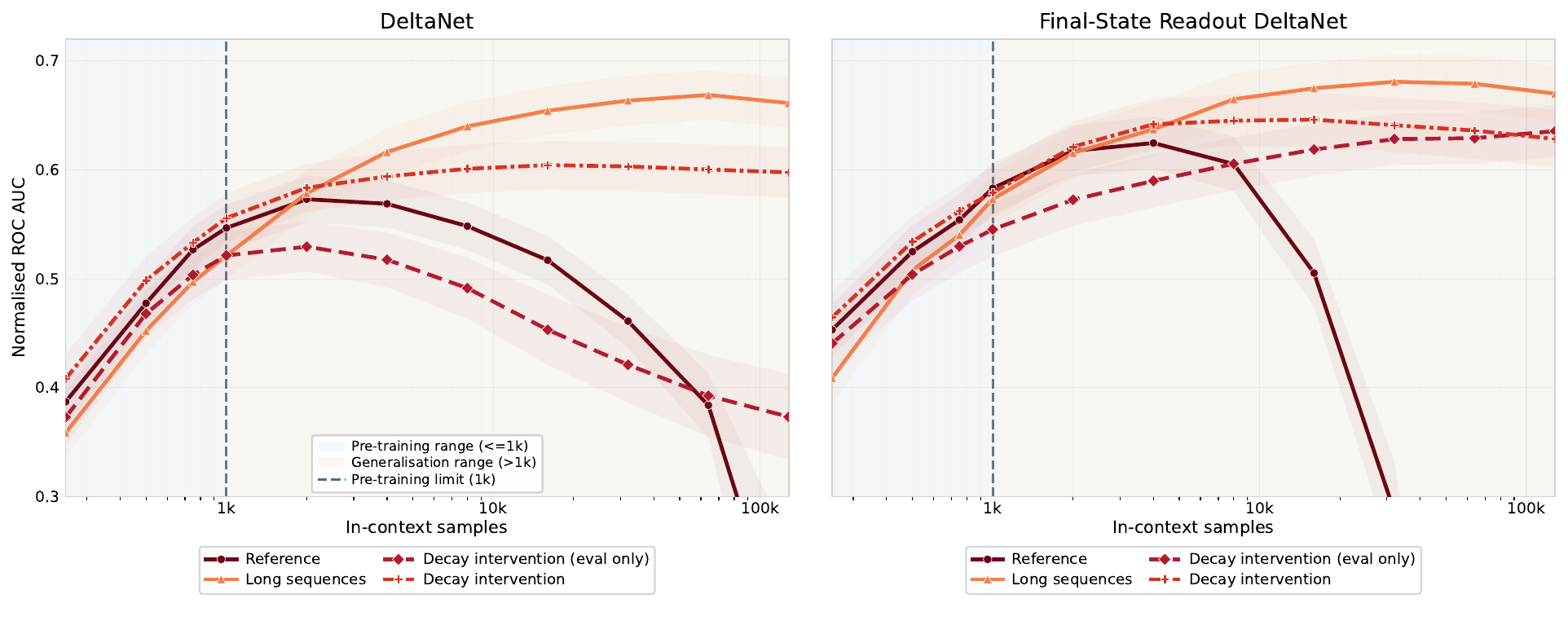}
    \caption{Length generalisation of standard DeltaNet (left) and final-state readout DeltaNet (right) under the decay intervention and long-sequence pretraining; \emph{eval only} applies the decay to the reference checkpoint at evaluation time. Long-sequence pretrained models are not compute-matched and serve only as analysis references.}
    \label{fig:standard_and_final_state_deltanet_with_decay_intervention}
\end{figure}

Figure~\ref{fig:standard_and_final_state_deltanet_with_decay_intervention} shows the resulting length generalisation of standard and final-state readout DeltaNet under the decay intervention of Section~\ref{subsec:recurrent_state-estimator} and under long-sequence pretraining.

\textbf{Final-state contribution analysis.} We additionally analyse a lower-bound proxy for the contribution of earlier positions to the final DeltaNet state. In the simplified case of identical unit-norm key directions, the contribution of a source position $s$ to the final state is
\begin{equation}
    C_s = \beta_s \prod_{t=s+1}^n (1-\beta_t).
\end{equation}

\begin{figure}[!h]
    \centering
    \includegraphics[width=\textwidth, trim={8pt, 20pt, 7pt 0pt}, clip]{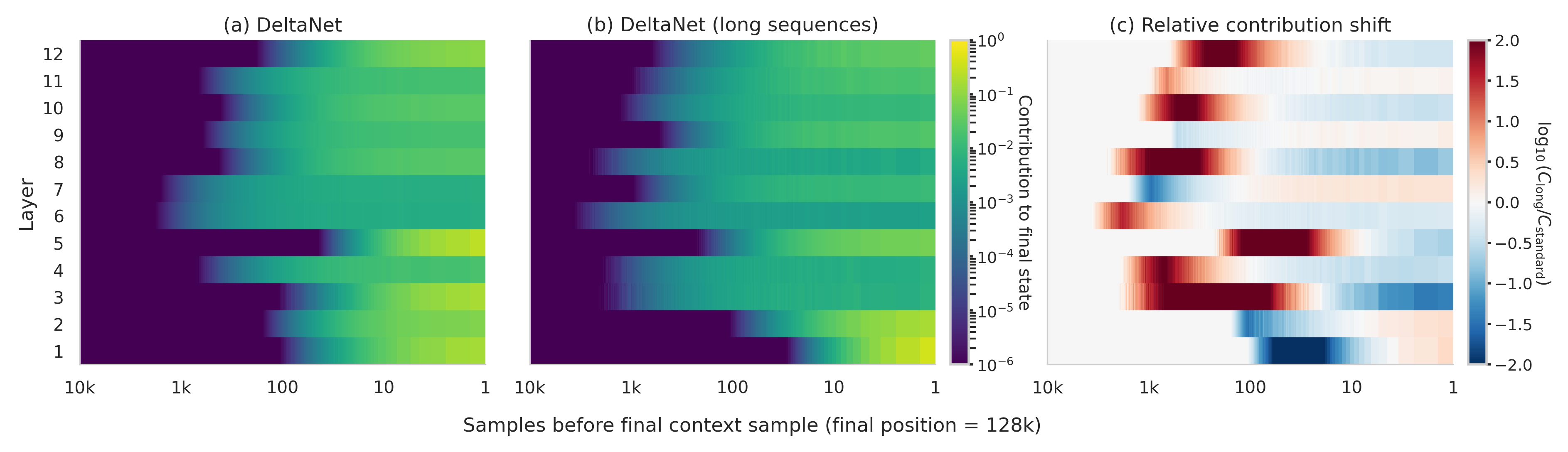}

    \includegraphics[width=\textwidth, trim={8pt, 0pt, 7pt 0pt}, clip]{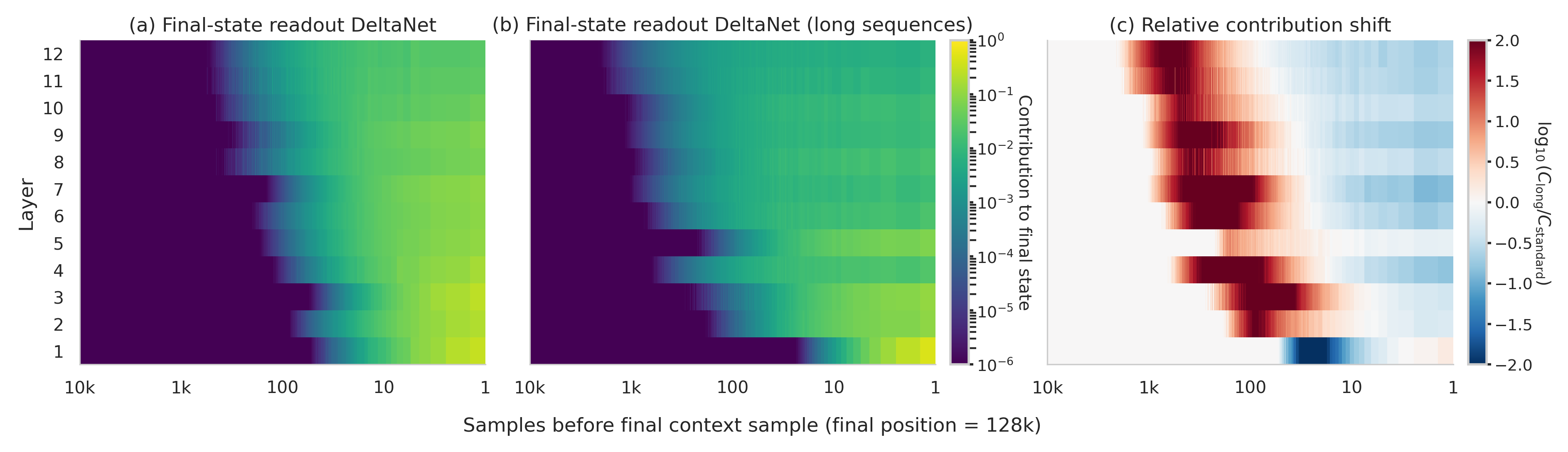}
    \caption{Contribution to the final DeltaNet hidden state across lookback distances. The top row compares DeltaNet with its long-sequence pretrained variant, and the bottom row compares the corresponding final-state readout variants. Panels (a,b) show a lower bound on the contribution $C_s = \beta_s \prod_{j>s}(1-\beta_j)$ of each source position $s$ to the final hidden state, averaged across 50 synthetic datasets and for each layer. Panel (c) shows the log-ratio $\log_{10}(C_{\mathrm{long}}/C_{\mathrm{standard}})$ where red indicates a larger contribution for the long-sequence pretrained model and blue indicates a larger contribution for the standard model.}
\label{fig:final_state_readout_DeltaNet_deltanet_final_state_contribution}
\end{figure}

Figure~\ref{fig:final_state_readout_DeltaNet_deltanet_final_state_contribution} shows these contributions across layers and lookback distances. Long-context pretraining variants retain more contribution from earlier samples, while the 1000-row pretrained variants mostly focus on recent samples. This supports the interpretation that DeltaNet overfits its write rates to the pretraining length range.

\newpage
\subsection{Extended Real-World Model Comparison}
\label{app:extended_model_comparison}

Tables~\ref{tab:extended_pairwise_openml} and~\ref{tab:extended_pairwise_tabarena} extend the comparison of Table~\ref{tab:model_comparison_real_world}. We here add the layer-local ridge state model, which is a closed-form solution to the layer-local objective of Equation~\ref{eq:ttr_objective} (Appendix~\ref{app:layer_local_ridge_details}), multiple subsampled variants of DeltaNet, and an Interleaved-Multi-Target (Int-MT) configuration of DeltaNet without decay. The 3K variants partition each context into disjoint blocks of at most 3000 rows and average the block logits.

\begin{table}[!ht]
\centering
\caption{Full comparison on the 30 OpenML-CC18$^*$ datasets. Models are ordered by ROC-AUC. FS denotes final-state readout, Decay denotes the time-dependent decay intervention, Subs.\ denotes the subsampled and ensembled configuration, and Int-MT denotes the Interleaved Multi-Target configuration (Appendix~\ref{app:Pre-training_and_Hyperparameter_details}). All other models use Combined Single-Target. Bold and underline mark the best and second-best mean per metric, while bold $p$-values mark adjusted $p\leq0.05$ indicating significant differences. The pairwise significance comparison uses two-sided Wilcoxon signed-rank tests with Holm adjustment like in Table~\ref{tab:model_comparison_real_world}. The 1K sample cap of OpenML-CC18$^*$ makes each Subs.\ 3K variant identical to its full-context counterpart.}
\label{tab:extended_pairwise_openml}
\vspace{0.8em}
\setlength{\tabcolsep}{0.75pt}
\footnotesize
\begin{tabular}{@{}rlrr*{11}{c}@{}}
\toprule
& & \multicolumn{2}{c}{Mean} & \multicolumn{11}{c}{Adjusted $p$ (ROC-AUC above diagonal, ACC below)} \\
\cmidrule(lr){3-4}\cmidrule(l){5-15}
\# & Model & AUC & ACC & 1 & 2 & 3 & 4 & 5 & 6 & 7 & 8 & 9 & 10 & 11 \\
\midrule
1 & Softmax Non-Causal & \textbf{0.8900} & \textbf{0.8166} & -- & 1.00 & .60 & 1.00 & 1.00 & 1.00 & .43 & .17 & .25 & .25 & .05 \\
2 & Delta-FS + Decay & \underline{0.8884} & \underline{0.8145} & 1.00 & -- & 1.00 & 1.00 & 1.00 & 1.00 & 1.00 & .21 & .27 & .27 & .77 \\
3 & Delta (Int-MT) & 0.8880 & 0.8098 & .51 & 1.00 & -- & 1.00 & 1.00 & 1.00 & 1.00 & 1.00 & .99 & .99 & 1.00 \\
4 & Delta-FS & 0.8874 & 0.8137 & 1.00 & 1.00 & \textbf{.04} & -- & 1.00 & 1.00 & 1.00 & 1.00 & .99 & .99 & 1.00 \\
5 & Delta-FS (Subs.\ 3K) & 0.8874 & 0.8137 & 1.00 & 1.00 & \textbf{.04} & 1.00 & -- & 1.00 & 1.00 & 1.00 & .99 & .99 & 1.00 \\
6 & Layer-Local Ridge & 0.8873 & 0.8131 & 1.00 & 1.00 & 1.00 & 1.00 & 1.00 & -- & 1.00 & .74 & 1.00 & 1.00 & 1.00 \\
7 & Delta + Decay (Int-MT) & 0.8873 & 0.8120 & .94 & 1.00 & 1.00 & .94 & .94 & 1.00 & -- & 1.00 & 1.00 & 1.00 & 1.00 \\
8 & Delta + Decay & 0.8864 & 0.8091 & .26 & .56 & 1.00 & .24 & .24 & .66 & 1.00 & -- & 1.00 & 1.00 & 1.00 \\
9 & Delta & 0.8860 & 0.8078 & .66 & .66 & 1.00 & \textbf{.04} & \textbf{.04} & 1.00 & 1.00 & 1.00 & -- & 1.00 & 1.00 \\
10 & Delta (Subs.\ 3K) & 0.8860 & 0.8078 & .66 & .66 & 1.00 & \textbf{.04} & \textbf{.04} & 1.00 & 1.00 & 1.00 & 1.00 & -- & 1.00 \\
11 & Linear Non-Causal & 0.8833 & 0.8039 & .17 & 1.00 & 1.00 & .10 & .10 & .28 & 1.00 & 1.00 & 1.00 & 1.00 & -- \\
\bottomrule
\end{tabular}

\end{table}

\begin{table}[!ht]
\centering
\caption{Full comparison on the 38 TabArena$^*$ datasets. Here we use the identical setup from Table~\ref{tab:extended_pairwise_openml}.}
\label{tab:extended_pairwise_tabarena}
\vspace{0.8em}
\setlength{\tabcolsep}{0.75pt}
\footnotesize
\begin{tabular}{@{}rlrr*{11}{c}@{}}
\toprule
& & \multicolumn{2}{c}{Mean} & \multicolumn{11}{c}{Adjusted $p$ (ROC-AUC above diagonal, ACC below)} \\
\cmidrule(lr){3-4}\cmidrule(l){5-15}
\# & Model & AUC & ACC & 1 & 2 & 3 & 4 & 5 & 6 & 7 & 8 & 9 & 10 & 11 \\
\midrule
1 & Softmax Non-Causal & \textbf{0.8172} & \textbf{0.8506} & -- & 1.00 & .79 & \textbf{.03} & \textbf{$<$.01} & \textbf{$<$.01} & \textbf{$<$.001} & \textbf{$<$.001} & \textbf{$<$.001} & .08 & \textbf{$<$.001} \\
2 & Delta + Decay (Int-MT) & \underline{0.8141} & 0.8459 & .74 & -- & 1.00 & .74 & \textbf{$<$.01} & \textbf{$<$.001} & \textbf{$<$.001} & \textbf{$<$.001} & \textbf{$<$.01} & .07 & \textbf{$<$.001} \\
3 & Delta-FS + Decay & 0.8139 & \underline{0.8468} & 1.00 & 1.00 & -- & \textbf{.01} & \textbf{$<$.01} & \textbf{$<$.001} & \textbf{$<$.001} & \textbf{$<$.001} & \textbf{$<$.001} & 1.00 & \textbf{$<$.001} \\
4 & Delta-FS (Subs.\ 3K) & 0.8123 & 0.8456 & 1.00 & 1.00 & .60 & -- & .55 & \textbf{.01} & \textbf{.01} & \textbf{$<$.001} & .43 & 1.00 & \textbf{$<$.001} \\
5 & Delta + Decay & 0.8100 & 0.8438 & .38 & .60 & .12 & 1.00 & -- & 1.00 & .11 & \textbf{.01} & 1.00 & 1.00 & \textbf{$<$.001} \\
6 & Delta (Subs.\ 3K) & 0.8090 & 0.8421 & \textbf{.01} & .09 & \textbf{$<$.01} & \textbf{.02} & .85 & -- & 1.00 & .51 & 1.00 & 1.00 & \textbf{$<$.01} \\
7 & Delta & 0.8074 & 0.8388 & \textbf{$<$.001} & \textbf{$<$.01} & \textbf{$<$.001} & \textbf{.01} & .42 & 1.00 & -- & 1.00 & 1.00 & 1.00 & \textbf{.03} \\
8 & Layer-Local Ridge & 0.8034 & 0.8418 & \textbf{$<$.01} & .38 & \textbf{.01} & .07 & .68 & 1.00 & 1.00 & -- & 1.00 & 1.00 & .08 \\
9 & Delta-FS & 0.8022 & 0.8352 & .71 & 1.00 & 1.00 & 1.00 & 1.00 & 1.00 & .35 & 1.00 & -- & 1.00 & .83 \\
10 & Delta (Int-MT) & 0.8004 & 0.8415 & .09 & 1.00 & .06 & 1.00 & 1.00 & 1.00 & 1.00 & 1.00 & 1.00 & -- & 1.00 \\
11 & Linear Non-Causal & 0.8004 & 0.8396 & \textbf{$<$.01} & \textbf{.02} & \textbf{$<$.001} & \textbf{.01} & .35 & .79 & 1.00 & .07 & .85 & 1.00 & -- \\
\bottomrule
\end{tabular}

\end{table}

\subsection{Two-Axis Row-Feature Processing}
\label{app:two_axis}

The main study uses row-wise processing, in which each row is compressed by a linear encoder into a single token. Modern tabular foundation models instead process two axes, attending across features as well as across rows \citep{hollmann2025tabpfn, qu2026tabiclv2}. We here test whether the length-generalisation failure holds for both causal DeltaNet and causal linear attention in a TabPFN-v2-style two-axis configuration. We use softmax attention across features and the recurrence across rows, under the same prior and protocol as the row-wise models.

Table~\ref{tab:two_axis_seq_len} reports mean ROC-AUC over the same 500 synthetic evaluation datasets. Two-axis processing does not resolve the failure, and its effect depends strongly on the update rule. For DeltaNet it increases the degradation, whereas causal linear attention remains comparatively stable. It is important to note that while we matched the number and size of the pretraining datasets, the two-axis models incur substantially higher computational cost in both memory and runtime per dataset and a higher parameter count of 14.9M due to the second axis, so they are not parameter-matched to the row-wise models.

\begin{table}[!h]
\centering
\setlength{\tabcolsep}{6pt}
\caption{Length generalisation under two-axis versus single-axis (row-wise) processing, reporting mean ROC-AUC over the 500 synthetic evaluation datasets by context length. Two-axis processing increases long-context degradation substantially for DeltaNet, while causal linear attention remains comparatively stable. The two-axis models are not parameter-matched to the row-wise models.}
\label{tab:two_axis_seq_len}
\vspace{0.5em}
\begin{tabular}{lcccc}
\toprule
Model & 1k & 8k & 32k & 128k \\
\midrule
DeltaNet, two-axis                  & 0.7808 & 0.7813 & 0.7592 & 0.7059 \\
DeltaNet, single-axis               & 0.7791 & 0.7806 & 0.7719 & 0.7569 \\
\midrule
Causal Linear Attention, two-axis    & 0.7777 & 0.7863 & 0.7852 & 0.7812 \\
Causal Linear Attention, single-axis & 0.7775 & 0.7805 & 0.7781 & 0.7723 \\
\bottomrule
\end{tabular}
\end{table}

\clearpage
\section{Memory and Compute Scaling Analysis}
\label{sec:scaling_analysis}

\begin{figure}[h]
    \centering
    \includegraphics[width=\textwidth]{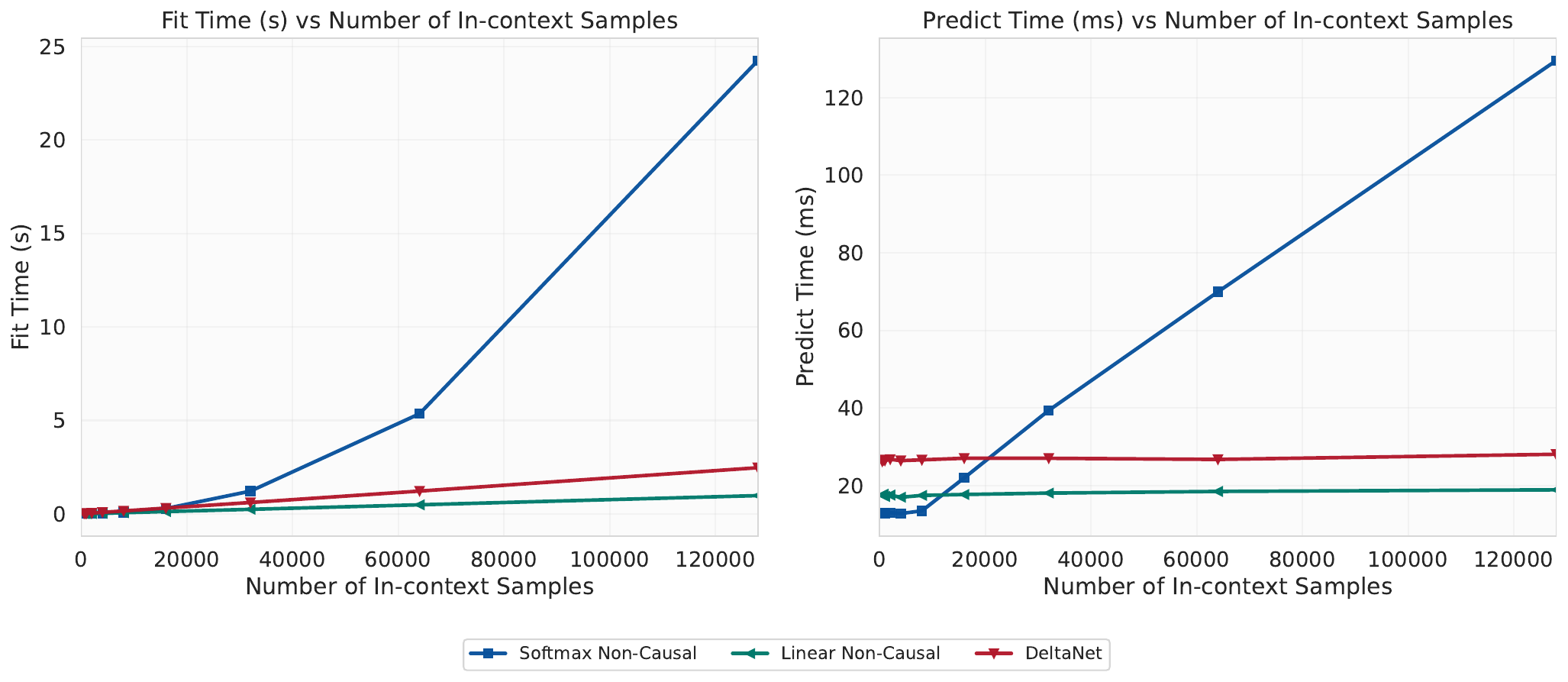}
    \caption{Runtime scaling with the number of in-context samples. Softmax attention scales quadratically during context processing and linearly during query processing with a cached context. Linear attention and DeltaNet scale linearly during context processing and remain constant during query processing from a fixed-size recurrent state.}
    \label{fig:compute_sacling_plot}
\end{figure}

\begin{figure}[h]
    \centering
    \includegraphics[width=.6\textwidth]{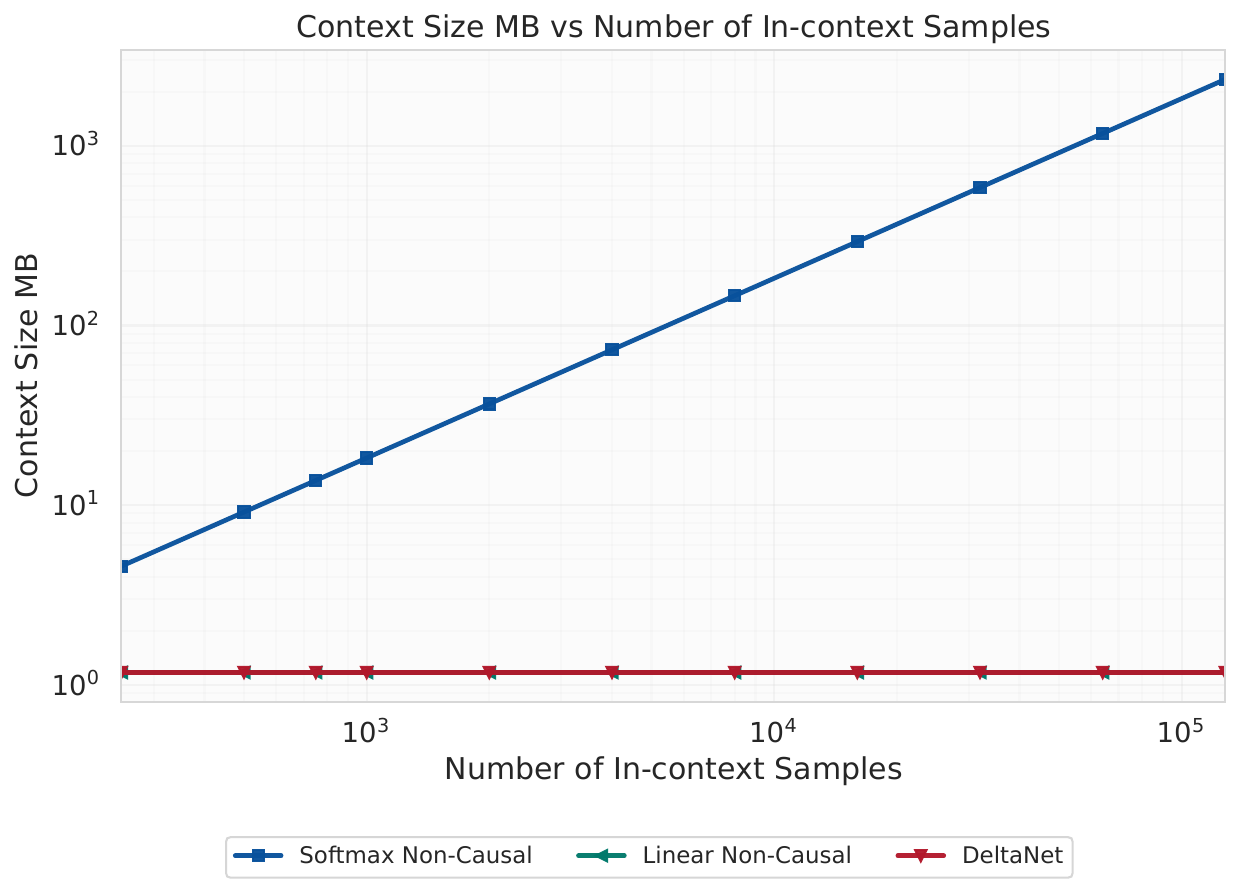}
    \caption{Context-state size across increasing context lengths. Softmax attention stores a KV cache that grows linearly with the number of context samples, whereas linear-time recurrent models store a fixed-size hidden state.}
    \label{fig:memory_scaling_plot}
\end{figure}

Figures~\ref{fig:compute_sacling_plot} and~\ref{fig:memory_scaling_plot} report the runtime and context-state size of the core context-processing and query-processing subroutines. Here, \emph{fit} denotes processing the labelled context and constructing the corresponding cache or recurrent state, not task-specific parameter training. \emph{Predict} denotes evaluating query samples given this cached context representation.

For softmax attention, context processing scales quadratically in the number of context samples, while query processing scales linearly when the context KV cache is already available. In contrast, non-causal linear attention and DeltaNet process the context in linear time and evaluate queries in constant time once the fixed-size recurrent state has been constructed. Similarly, the softmax KV cache grows linearly with the number of context samples, whereas recurrent models store a fixed-size hidden state.

The measurements reported in the figures correspond to a single preprocessing configuration. In real-world tabular foundation-model inference, predictions are typically ensembled across multiple preprocessing configurations. For TabPFN v1 and v2, this is referred to as an ensemble configuration \citep{hollmann2023tabpfn,hollmann2025tabpfn}, often using $N_{\mathrm{ens}} \in \{10, \ldots, 32\}$ transformed versions of the same dataset. Algorithm~\ref{alg:prediction_with_preprocessing_ensembles} shows this procedure.

\begin{algorithm}[!h]
\caption{Prediction with preprocessing ensembles}
\label{alg:prediction_with_preprocessing_ensembles}
\begin{algorithmic}[1]
\Require Context $(X_{\mathrm{ctx}}, y_{\mathrm{ctx}})$, queries $X_{\mathrm{qry}}$, preprocessing configurations $\{\pi_e\}_{e=1}^{N_{\mathrm{ens}}}$
\For{$e = 1,\dots,N_{\mathrm{ens}}$}
    \State $(\tilde X_{\mathrm{ctx}}^{(e)}, \tilde y_{\mathrm{ctx}}^{(e)}, \tilde X_{\mathrm{qry}}^{(e)}) \gets \pi_e(X_{\mathrm{ctx}}, y_{\mathrm{ctx}}, X_{\mathrm{qry}})$
    \State $C^{(e)} \gets \mathrm{Forward}_{\mathrm{ctx}}(\tilde X_{\mathrm{ctx}}^{(e)}, \tilde y_{\mathrm{ctx}}^{(e)})$
    \State $\hat z^{(e)} \gets \mathrm{Forward}_{\mathrm{qry}}(C^{(e)}, \tilde X_{\mathrm{qry}}^{(e)})$
\EndFor
\State $\bar z \gets \frac{1}{N_{\mathrm{ens}}}\sum_{e=1}^{N_{\mathrm{ens}}}\hat z^{(e)}$
\State \Return $\mathrm{softmax}(\bar z)$
\end{algorithmic}
\end{algorithm}

Naïvely, the context and queries are processed together, resulting in a combined fit and predict time of
\begin{equation}
    T_{\mathrm{qry,total}} = N_{\mathrm{ens}} \cdot (T_{\mathrm{ctx}} + T_{\mathrm{qry}}), 
\end{equation}
where $T_{\mathrm{ctx}}$ denotes the \emph{fit} time required to process the context, and $T_{\mathrm{qry}}$ denotes the \emph{predict} time required to process the queries given the processed context. However, for every additional query, the full training context state would need to be recomputed again. 

Alternatively, the fit and predict stage can be separated and amortised, but this requires caching the context representations (KV-cache or hidden-state $\bm{S}_{n_c}$) for all ensemble members. In this algorithm,
the effective costs scale as
\begin{align}
    T_{\mathrm{ctx,total}} &= N_{\mathrm{ens}} \cdot T_{\mathrm{ctx}}, \\
    T_{\mathrm{qry,total}} &= N_{\mathrm{ens}} \cdot (T_{\mathrm{qry}} + T_{\mathrm{load}}), \\
    M_{\mathrm{cache,total}} &= N_{\mathrm{ens}} \cdot M_{\mathrm{cache}},
\end{align}
where $M_{\mathrm{cache}}$ denotes the size of the context representation and $T_{\mathrm{load}}$ denotes the time required to load the cached context representation for one ensemble member when the cache needs to be offloaded from GPU memory to main memory or disk. The implementation of such an ensemble caching algorithm is beyond the scope of this project. We therefore omit a separated end-to-end fit and predict runtime analysis and report the pure $T_{\mathrm{ctx}}$, $T_{\mathrm{qry}}$, and $M_{\mathrm{cache}}$ in Figures~\ref{fig:compute_sacling_plot} and \ref{fig:memory_scaling_plot} instead.

This ensemble effect further highlights the memory advantage of parametric tabular foundation models. In our setting, the softmax-transformer KV cache at sequence length 128k requires 
2.34\,GB,
whereas the DeltaNet recurrent state requires only 1.17\,MB. With $N_{\mathrm{ens}}=10$, this corresponds to 23.4\,GB for softmax attention compared to 11.7\,MB for DeltaNet. For datasets with millions of samples, the KV cache would grow significantly, limiting real-world applicability both in runtime and memory cost. Additionally, it would likely require disk offloading of the cache, further increasing inference latency.

The memory burden further increases along the feature axis under two-axis processing (Appendix~\ref{app:two_axis}), where both the cache and the recurrent state scale with the number of feature groups. A dataset with 100 features and a feature grouping of five results in 20 groups, multiplying the memory requirements of both approaches by $20$. A 10-member ensemble would then cache roughly 469\,GB of keys and values against 234\,MB of recurrent state. Such small states, combined with constant-time query processing, could allow linear-time tabular foundation models to be deployed on edge devices.

\section{Adding Stateless Prediction to Flash-Linear-Attention Models}
\label{app:adding_stateless_prediction_to_fla}

Causal linear attention architectures used in our experiments are imported from the FLA library~\citep{yang2024fla}. These implementations natively support autoregressive sequence processing. However, as discussed in Section~\ref{subsec:Training_strategies}, non-transductive evaluation requires a different prediction mode: each query sample must be predicted independently from the context set, without conditioning on any other query samples. We refer to this setting as \emph{single-target} prediction in the main paper.

We considered three approaches for adding this prediction mode:
\begin{enumerate}
    \item \textbf{Naive implementation:} 
    The simplest solution is to evaluate each query sample separately. For a test set with $n_q$ query samples, this requires $n_q$ forward passes. One can either run the full sequence consisting of the context set and one query sample each time or first process the context set once, cache the resulting recurrent state, and then perform $n_q$ single-sample forward passes while resetting the recurrent state to the cached context state after each query. The latter avoids repeatedly processing the context, but the query evaluations remain sequential.
    
    \item \textbf{Broadcasting the recurrent state:} 
    A natural way to parallelise the naive approach is to broadcast the final context recurrent state along a new query dimension or, equivalently, expand the batch dimension by a factor of $n_q$. Each query could then be evaluated in parallel from the same context state. However, the FLA implementations materialise the expanded recurrent state rather than representing it as virtual views. This leads to $n_q$ explicit copies of the final recurrent states. For query sets with hundreds or thousands of samples, the resulting memory cost is prohibitive, and the additional computation is inefficient.

    \item \textbf{Custom stateless patch:} 
    We therefore implement a custom stateless prediction patch. The goal is to obtain the parallel one-step prediction behaviour of the broadcasted implementation without materialising $n_q$ copies of the recurrent state. Since single-target evaluation only requires the prediction for each query sample, and not the recurrent state after the query has been processed, the query update can be specialised. We use the standard FLA kernels to process the context samples, while during query prediction, we reuse the FLA model logic but replace the recurrent kernel update with a custom PyTorch implementation that applies the cached context state independently to all query samples without writing back query-dependent states.
\end{enumerate}

To illustrate the \emph{custom stateless patch} that we use in this paper, Algorithm~\ref{alg:stateless_linear_attention} shows the corresponding stateless one-step computation for a linear attention layer with an additive recurrent state update. For readability, we omit the standard query scaling factor.

\begin{algorithm}[!ht]
\begin{algorithmic}[1]
\Require Cached context state $S_c \in \mathbb{R}^{B \times H \times d_v \times d_{qk}}$
\Require Query projections $q,k \in \mathbb{R}^{B \times n_q \times H \times d_{qk}}$ and $v \in \mathbb{R}^{B \times n_q \times H \times d_v}$

\State $O^{\mathrm{ctx}} \gets \mathrm{einsum}(\texttt{"bihk,bhvk->bihv"}, q, S_c)$
\Comment{shared context readout}

\State $s \gets \mathrm{einsum}(\texttt{"bihk,bihk->bih"}, q, k)$
\Comment{query-key scores}

\State $O^{\mathrm{self}} \gets \mathrm{einsum}(\texttt{"bih,bihv->bihv"}, s, v)$
\Comment{self terms}

\State $O \gets O^{\mathrm{ctx}} + O^{\mathrm{self}}$
\Comment{final output}

\State \Return $O$
\end{algorithmic}
\caption{Stateless single-target prediction for one linear attention layer}
\label{alg:stateless_linear_attention}
\end{algorithm}

For a standard linear-attention recurrent layer, the state update and readout are
\begin{equation}
    \bm{S}_t = \bm{S}_{t-1} + \bm{v}_t \bm{k}_t^\top,
    \qquad
    \bm{o}_t = \bm{S}_t \bm{q}_t.
\end{equation}
After processing the context samples, we obtain a cached context state $\bm{S}_c$. For an independent query sample $i$, the standard one-step computation would first form the query-updated state
\begin{equation}
    \bm{S}_i = \bm{S}_c + \bm{v}_i \bm{k}_i^\top,
    \qquad
    \bm{o}_i = \bm{S}_i \bm{q}_i.
\end{equation}
Expanding the readout gives
\begin{equation}
\begin{aligned}
    \bm{o}_i
    &= \bm{S}_i \bm{q}_i \\
    &= (\bm{S}_c + \bm{v}_i \bm{k}_i^\top) \bm{q}_i \\
    &= \bm{S}_c \bm{q}_i + (\bm{v}_i \bm{k}_i^\top) \bm{q}_i \\
    &= \bm{S}_c \bm{q}_i + (\bm{k}_i^\top \bm{q}_i)\bm{v}_i .
\end{aligned}
\end{equation}
The stateless patch computes this expanded form directly for all query samples in parallel. Thus, it reads from the shared context state $\bm{S}_c$ and adds each query's own one-step contribution, without materialising the query-updated states $\bm{S}_i$.

\end{document}